\documentclass{article}

\usepackage{yfonts}

\usepackage{microtype}
\usepackage{graphicx}
\usepackage{subfigure}
\usepackage{booktabs} 

\usepackage{hyperref}

\usepackage[accepted]{icml2023}

\usepackage{amsmath}
\usepackage{amssymb}
\usepackage{mathtools}
\usepackage{amsthm}
\usepackage[most]{tcolorbox}

\usepackage{tabularx}

\newcolumntype{M}[1]{>{\centering\arraybackslash}m{#1}} 
\newcolumntype{R}[1]{>{\raggedleft\arraybackslash}m{#1}} 
\newcolumntype{L}[1]{>{\raggedright\arraybackslash}m{#1}} 
\usepackage{booktabs} 
\usepackage{multirow}

\usepackage[capitalize,noabbrev]{cleveref}

\theoremstyle{plain}

\theoremstyle{definition}

\theoremstyle{remark}

\usepackage[textsize=tiny]{todonotes}

\icmltitlerunning{PipeRAG: Fast Retrieval-Augmented Generation via Algorithm-System Co-design}

\begin{document}

\twocolumn[
\icmltitle{\large PipeRAG: Fast Retrieval-Augmented Generation via Algorithm-System Co-design}




\begin{icmlauthorlist}
\icmlauthor{Wenqi Jiang$^\dagger$}{eth}
\icmlauthor{Shuai Zhang}{aws}
\icmlauthor{Boran Han}{aws}
\icmlauthor{Jie Wang$^\dagger$}{meta}
\icmlauthor{Bernie Wang}{aws}
\icmlauthor{Tim Kraska}{aws,mit}
\end{icmlauthorlist}

\icmlaffiliation{eth}{ETH Zurich}
\icmlaffiliation{aws}{Amazon Web Services}
\icmlaffiliation{meta}{Meta}
\icmlaffiliation{mit}{MIT}

\icmlcorrespondingauthor{Wenqi Jiang}{wenqi.jiang@inf.ethz.ch}
\icmlcorrespondingauthor{Shuai Zhang}{shuaizs@amazon.com}

\icmlkeywords{Machine Learning, ICML}

\vskip 0.3in
]



\printAffiliationsAndNotice{}  
\def\thefootnote{$^\dagger$}\footnotetext{Work conducted while at Amazon.}\def\thefootnote{\arabic{footnote}}
\begin{abstract}
Retrieval-augmented generation (RAG) can enhance the generation quality of large language models (LLMs) by incorporating external token databases. However, retrievals from large databases can constitute a substantial portion of the overall generation time, particularly when retrievals are periodically performed to align the retrieved content with the latest states of generation.
In this paper, we introduce PipeRAG, a novel algorithm-system co-design approach to reduce generation latency and enhance generation quality.
PipeRAG integrates (1) pipeline parallelism to enable concurrent retrieval and generation processes, (2) flexible retrieval intervals to maximize the efficiency of pipeline parallelism, and (3) a performance model to automatically balance retrieval quality and latency based on the generation states and underlying hardware.
Our evaluation shows that, by combining the three aforementioned methods, PipeRAG achieves up to 2.6$\times$ speedup in end-to-end generation latency while improving generation quality. These promising results showcase the effectiveness of co-designing algorithms with underlying systems, paving the way for the adoption of PipeRAG in future RAG systems.


\end{abstract}

\section{Introduction}
\label{sec:intro}


Retrieval-augmented generation (RAG) enhances auto-regressive large language models (LLMs) by conditioning on contextually relevant content retrieved from external databases. 
While one retrieval prior to the generation process can be enough when generating short sequences~\cite{lewis2020retrieval, izacard2020leveraging}, a more general approach involves periodic retrievals throughout the generation~\cite{borgeaud2022improving, norlund2023generalization, ram2023context, jiang2023active, trivedi2022interleaving}. This necessity arises due to the potential shift in the generation context, such as changes in topics.  Therefore, periodic retrievals ensure the retrieved content remains relevant to the latest context of the generation (Appendix~\ref{sec:append_more_background} showcases a concrete example). 
A popular example of this category is \textsc{Retro}~\cite{borgeaud2022improving}, which tailors the transformer neural network architecture to support the integration of retrieved content at regular intervals.

However, periodic retrievals on large databases, potentially comprising trillions of tokens~\cite{borgeaud2022improving}, can lead to a significant slowdown of the sequence generation. \textit{We ask: can we optimize the system performance of RAG while preserving or even improving generation quality?} 

\begin{figure*}[t]
	\centering
  \includegraphics[width=0.87\linewidth]{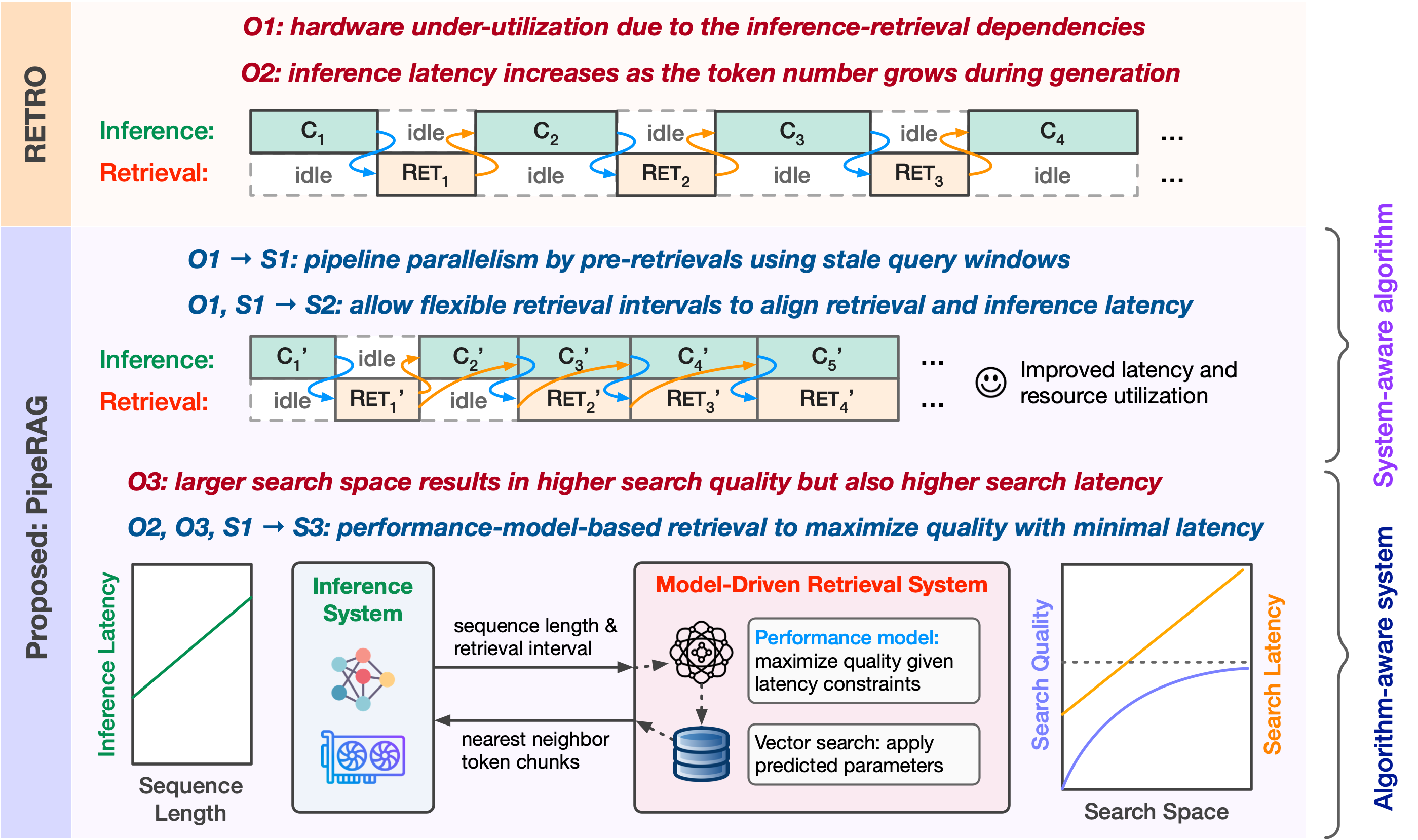}
  \vspace{-1.2em}
  \caption{Based on three performance-centric observations (O1$\sim$O3), PipeRAG combines a system-aware algorithm integrating pipeline parallelism (S1) with flexible retrieval intervals (S2) and an algorithm-aware retrieval system guided by a performance model (S3).}
  \vspace{-1.2em}
  \label{fig:overview}
\end{figure*}

 
We propose PipeRAG, a pioneering approach to improve RAG efficiency via a collaborative algorithm-system co-design --- including a system-aware RAG algorithm and an algorithm-aware retrieval system as overviewed in Figure~\ref{fig:overview}. 

The foundation of PipeRAG is established on three observations centered on performance. Firstly, the dependencies between retrievals and LLM inferences lead to hardware underutilization, with either the inference or retrieval system being idle at any given time during the generation process~(\textbf{O1}). Secondly, the inference latency per token increases with sequence lengths, due to the growing workloads of the attention mechanism in transformer neural networks~(\textbf{O2}). Lastly, the retrieval process, particularly the approximate nearest neighbor search, exhibits a trade-off between search latency and search quality~(\textbf{O3}).

\textit{The key idea of PipeRAG is to prefetch content from databases to facilitate pipeline parallelism between the inference and retrieval systems.} This solution reduces end-to-end generation latency by allowing simultaneous inference and retrievals, effectively addressing the hardware inefficiencies identified in O1~(\textbf{S1}). 
We then enhance this key idea with two additional solutions.
On the model side, PipeRAG modifies \textsc{Retro}'s attention mechanism to support flexible retrieval intervals, because the intervals must be carefully tuned to capitalize the efficiency of pipeline parallelism~(\textbf{S2}). 
On the system side, the retrieval system adopts a performance model informed by O2 and O3 to dynamically adjust the retrieval search space according to the latency expectation of the upcoming token inferences in the pipeline, thereby optimizing search quality without increasing end-to-end generation latency~(\textbf{S3}).

Our evaluation of PipeRAG, involving various evaluation datasets and using large databases based with up to 200 billion tokens, clearly illustrates its efficiency in both generation performance (latency) and generation quality (perplexity). Specifically, the quality-performance Pareto frontier of PipeRAG significantly outperforms that of \textsc{Retro}: PipeRAG can achieve up to 2.6$\times$ speedup in latency without compromising perplexity; alternatively, maintaining the same latency allows PipeRAG to reduce perplexity by as much as 0.93 compared to \textsc{Retro}. These encouraging results highlight the importance of algorithm-system co-design in retrieval-augmented generation, paving the way for deploying PipeRAG in future RAG systems.

\textbf{Contributions:} We propose PipeRAG, the first algorithm-system co-design approach aimed at improving retrieval-augmented generation efficiency. Specifically:
\begin{itemize}
\vspace{-0.9em}
\item We design a system-aware RAG algorithm that leverages pipeline parallelism, whose efficiency is further improved by supporting flexible retrieval intervals.
\vspace{-0.5em}
\item We propose an algorithm-aware retrieval system that uses performance models to dynamically balance search quality and performance.
\vspace{-0.5em}
\item We showcase the impressive performance of PipeRAG in various datasets, demonstrating the importance of algorithm-system co-design in optimizing RAG.
\end{itemize}

\section{Background and Motivation}
\label{sec:background}



\begin{figure}
	\centering
  \includegraphics[width=0.9\linewidth]{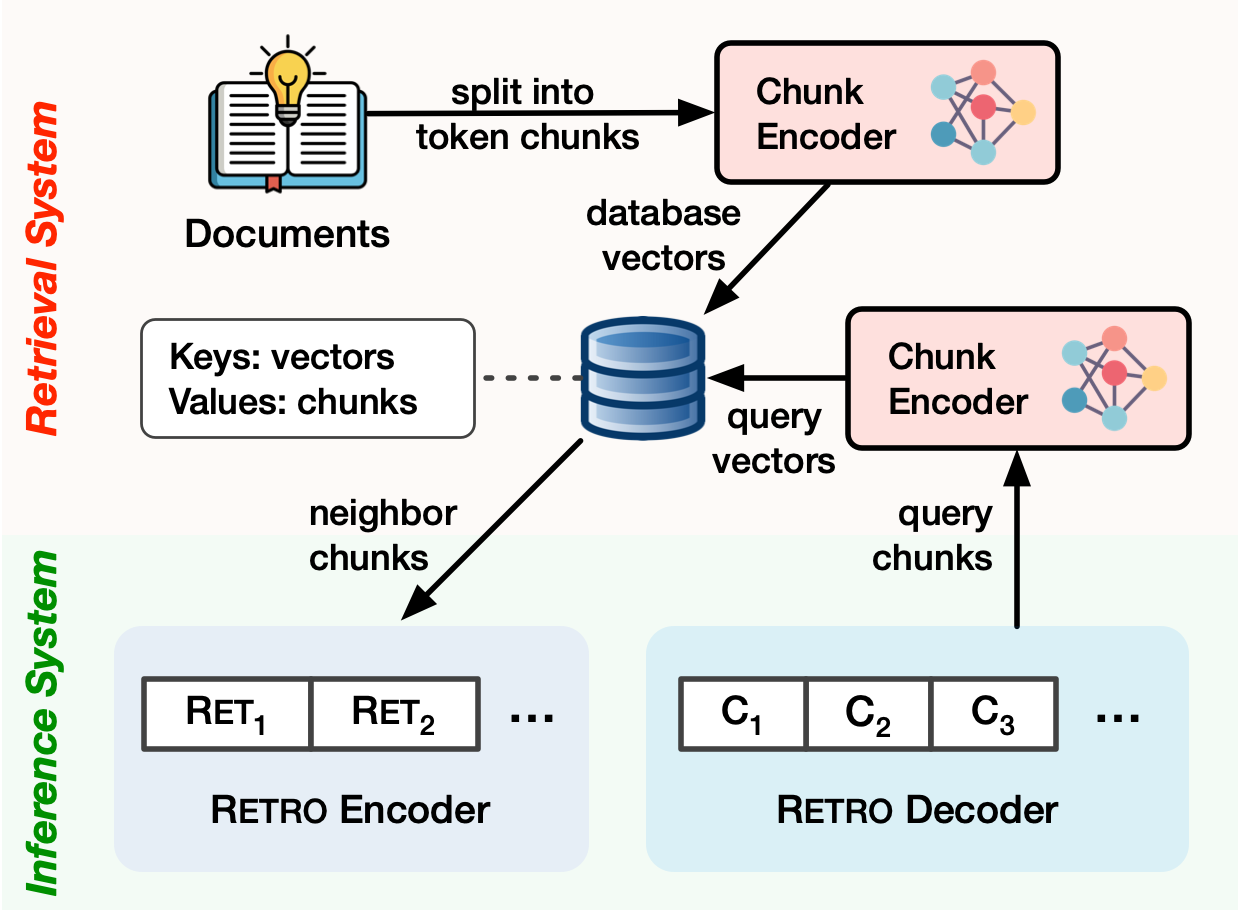}
  \vspace{-1em}
  \caption{Retrieval-augmented generation with \textsc{Retro}.}
  \vspace{-2em}
  \label{fig:retro}
\end{figure}

Sequence generation quality of LLMs can be improved through periodically retrieving from large token databases~\cite{borgeaud2022improving, norlund2023generalization, ram2023context}. Here, periodic retrievals, instead of retrieving only once, are essential in handling potential contextual shifts during generation, such as topic changes, ensuring alignments between the retrieved content and the evolving generation context (a concrete example can be found in Appendix~\ref{sec:append_more_background}). 
\textsc{Retro} is a representative model in this category~\cite{borgeaud2022improving}. 
As illustrated in Figure~\ref{fig:retro}, \textsc{Retro} integrates a retrieval system with an inference system for token generation. It employs an encoder for incorporating retrieved tokens and a decoder for token generation.

\textbf{Database construction.} A \textsc{Retro} database comprises a large collection of documents segmented into \( n \) chunks of tokens \( S = (S_1, \ldots, S_n) \), where each chunk \( S_i \) spans \( m \) tokens. These token chunks are each converted into vector representations \( R(S) \). The database is then structured as a key-value store, with keys being the vector representations \( R(S) \) and values corresponding to the original token chunks \( S \), along with \( F \), in which \( F_i \) representing the immediately following token chunks of each chunk \( S_i \). Given a query vector \( q \), the database performs an approximate nearest neighbor (ANN) search to retrieve \( k \) closest token chunks and their immediately following chunks. 

\textbf{Retrieval process.}  \textsc{Retro} performs retrievals at regular intervals during the generation phase. Specifically, when generating a sequence of $t$ tokens \( X = (x_1, \ldots, x_t) \), \textsc{Retro} partitions \( X \) into \( l \) chunks \( (C_1, \ldots, C_l) \), each consisting of \( m \) tokens. Consequently, token \( x_{i} \) belongs to chunk \( C_{\lceil \frac{i}{m} \rceil} \). For the generation of chunk \( C_i \), \textsc{Retro} employs the preceding chunk \( C_{i-1} \) as the query to retrieve \( k \) nearest neighbors \( \textsc{Ret}(C_{i-1}) \) from the database.

\textbf{Attention mechanisms.} 
\textsc{Retro} involves both decoder-to-encoder and encoder-to-decoder attention mechanisms. 
The decoder within \textsc{Retro} utilizes chunked cross-attention to integrate the retrieved information encoded by the encoder. To preserve causality, the generation of a chunk $C_i$ incorporates the retrieved tokens \( \textsc{Ret}(C_{i-1})\) by integrating the encoder states \( \textsc{Enc}(\textsc{Ret}(C_{i-1}))\).
On the other hand, the \textsc{Retro} encoder states \( \textsc{Enc}(\textsc{Ret}(C_{i-1}))\) integrates the decoder's states of the \( \textsc{Dec}(C_{i-1}) \) via a standard cross-attention (CA) mechanism, such that the encoder can blend the retrieved information with the generation context.
Because both decoder-to-encoder and encoder-to-decoder attention mechanisms operate on a chunk-wise basis, \textsc{Retro} avoids the excessive computational demands of attending to all previous retrieval and generation states.

\textbf{Motivation: improving RAG efficiency.}
Although periodically retrieving tokens from a large database can effectively improve the generation quality of LLMs, frequent retrievals can account for a considerable portion of the total generation time, thereby significantly slowing down the end-to-end generation process.

In this paper, we ask the following question: \textbf{is it possible to further enhance the efficiency of retrieval augmented generation?} Here, we conceptualize \textit{RAG efficiency} as a Pareto frontier considering two objectives: \textit{generation quality} and \textit{system performance}. 
Specifically, given a quality requirement (achieving certain perplexity), can we optimize RAG's system performance (reducing generation latency)? On the other hand, given a system performance requirement, can we improve the quality of generation?

\section{Our Approach: PipeRAG}
\label{sec:approach}

We propose PipeRAG, a novel retrieval augmented generation approach to improve the performance-quality Pareto frontier through an in-depth algorithm-system co-design.
The development of PipeRAG stems from performance-centric observations revealing (1) the \textit{fundamental} system inefficiencies in existing RAG algorithms and (2) the distinct performance characteristics of LLM inference and retrieval systems. 
Based on these observations, PipeRAG includes (1) a system-aware RAG algorithm to address the system inefficiencies and (2) an algorithm-aware retrieval system to dynamically balance retrieval quality and latency. 

\subsection{Performance-Centric Observations in RAG}
 
\textbf{O1: Hardware inefficiency due to RAG dependencies.} A conventional RAG process introduces dependencies between retrievals and inferences: the current generation context is used as a query to retrieve relevant token chunks stored in the database; the inference process must wait for the retrieval to finish before it can continue generating a few more tokens, until the next retrieval is triggered.

A RAG system typically comprises two sub-systems: the retrieval system and the inference system, each hosted on separate hardware platforms.
AI accelerators such as GPUs and TPUs are the ideal hardware platforms for LLM inference due to the high demands for computation and memory bandwidth during inference. 
On the other hand, the retrieval systems consisting of large databases are usually not based on GPUs. This is because (1) the limited memory capacity of individual GPUs (GPUs adopt high-bandwidth memory that is fast but limited in capacity) makes the hosting of large databases cost-prohibitive, necessitating the setup comprising many GPUs, and (2) the communication bandwidth between the CPU and GPU is significantly lower compared to GPU's device memory bandwidth, thus the CPU-GPU solution, in which database vectors are stored in CPU-side memory and then transferred to GPUs at query time, could be exceedingly slow.
Given the capacity requirements, the retrieval system is typically CPU-based~\cite{borgeaud2022improving, lewis2020retrieval}, with the database either held in substantial main memory (DRAM), or, in more budget-friendly setups, stored on disks. 

Given that the two systems are based on separate hardware, the dependencies between retrievals and inferences in RAG result in significant underutilization of hardware resources. Figure~\ref{fig:overview} illustrates this inefficiency using RETRO as a representative example: due to the dependencies, either the inference or retrieval system is idle at any given time during the generation process, leading to hardware inefficiencies.

\textbf{O2: Increasing inference time with sequence length.} 
In a standard transformer neural network~\cite{vaswani2017attention}, the cost of generating each new token correlates with the sequence length, rather than remaining a constant. This is due to the attention mechanism in transformers: although the workload of the fully-connected layers remains constant throughout the generation process, the cost of attention layers increases with the sequence length~\cite{beltagy2020longformer}. Specifically, for each new token generated, the query states (Q) of the most recent token are compared against the key states (K) of all preceding tokens to calculate relevance scores. These scores are then utilized for a weighted sum over the value states (V) (note that the queries, keys, and values mentioned here under the context of transformers are distinct from those terms in RAG systems).
Consequently, the inference cost per token can be approximated as a linear function to sequence length.

\textbf{O3: Trade-offs between retrieval quality and latency.} Large-scale vector search in RAG employs approximate nearest neighbor (ANN) search instead of exact nearest neighbor search due to the latter's prohibitive cost on large databases. In ANN search, database vectors are indexed, with popular choices including clustering-based inverted-file (IVF) indexes~\cite{IVF} and graph-based indexes~\cite{malkov2014approximate, malkov2018efficient}. Optionally, database vectors may also be compressed via product quantization (PQ)~\cite{PQ} to shrink database sizes and reduce memory bandwidth usage at query time at the expense of search accuracy. During a search, a query vector is only compared against a subset of database vectors selected by the index. 

Regardless of the index types, there exists a \textit{fundamental} trade-off between search quality and latency in ANN search. Typically, the index first directs the search towards those database vectors that are most likely to be the nearest neighbors of the query vector, and then gradually expands the search space. The number of database vectors scanned per query can be directly or indirectly controlled by ANN search hyper-parameters. Expanding the search space would enhance the probability of finding the query vector's true nearest neighbors in the database (improved search quality), but also would also lead to higher latency (lower search performance) due to the greater number of comparisons between query vectors and database vectors.

Figure~\ref{fig:overview} visualizes the relationship between search quality and latency~\cite{PQ}. As the search space expands (number of scanned database vectors), the search quality (recall of the retrieval) gradually improves until reaching a plateau where the nearest neighbors are likely found. Simultaneously, the search cost (latency) grows linearly with the search space, with an initial cost of scanning the index (which could be zero in some graph-based indexes). 

\subsection{Algorithm-System Co-deisgn in PipeRAG}

Given the aforementioned performance-centric observations, we propose PipeRAG, an algorithm-system co-design approach aimed at enhancing RAG's performance-quality Pareto frontier. 
PipeRAG addresses the \textit{fundamental} issue of hardware inefficiency (O1) by employing pipeline parallelism (S1) and allowing flexible retrieval intervals (S2). Leveraging the distinct performance characteristics of the inference and retrieval sub-systems (O2, O3), PipeRAG further offers an option to enable automatic search space selection within the retrieval system, facilitating high-quality generation without introducing additional generation latency.

\textbf{S1: Pipeline parallelism across RAG sub-systems.} 
Because the hardware under-utilization issue in RAG is caused by dependencies between retrievals and inferences, our first solution is about revisiting RAG algorithms to enable pipeline parallelism: the retrievals and inferences should be executed concurrently, thus overlapping their execution latency and improving hardware utilization. 

To facilitate pipeline parallelism, we relax the RAG dependencies as illustrated in Figure~\ref{fig:overview}: instead of depending on the content retrieved using the query representing the most recent generation context (the latest generated tokens),
the inference process can utilize a slightly older, or \textit{stale}, query window to \textit{prefetch} content from the database.
The intuition here is that if the stale query window closely aligns with the latest generation context, it is likely to retrieve content similar to that obtained using the most recent query tokens.
Once the dependency constraint is relaxed, retrievals can be proactively initiated to \textit{prefetch} content from the database, thus enabling pipeline parallelism as shown in Figure~\ref{fig:overview}.

Formally, when generating token chunk \( C_{j+1} \), PipeRAG does not use the immediately preceding chunk as the query \( Q = C_j = (x_{jm}, \ldots, x_{jm + m - 1}) \) to retrieve \( \textsc{Ret}(Q) \). Instead, it opts for a stale token window \( \hat{Q} = (x_{jm - s}, \ldots, x_{jm + m - 1 - s}) \) as an approximate query, offset by \( s \) tokens from the latest query window. Subsequently, \( \hat{\textsc{Ret}(Q)} = \textsc{Shift}(\textsc{Ret}(\hat{Q}), s) \) serves as the approximation of \( \textsc{Ret}(Q) \). Given that the stale query is \( s \) tokens behind the most recent generation context, the retrieved results \( \textsc{Ret}(\hat{Q}) \) are correspondingly left-shifted by \( s \) tokens. This shift ensures that the first \( s \) retrieved tokens, which are likely less relevant for the upcoming generation due to staleness, are excluded while maintaining the overall length of retrieval tokens. Note that the concept of stale query windows does not apply for the initial retrieval, which is conducted using the first chunk \( C_{1} \), as illustrated in Figure~\ref{fig:overview}.

\begin{figure}[t]
	\centering
  \includegraphics[width=1.0\linewidth]{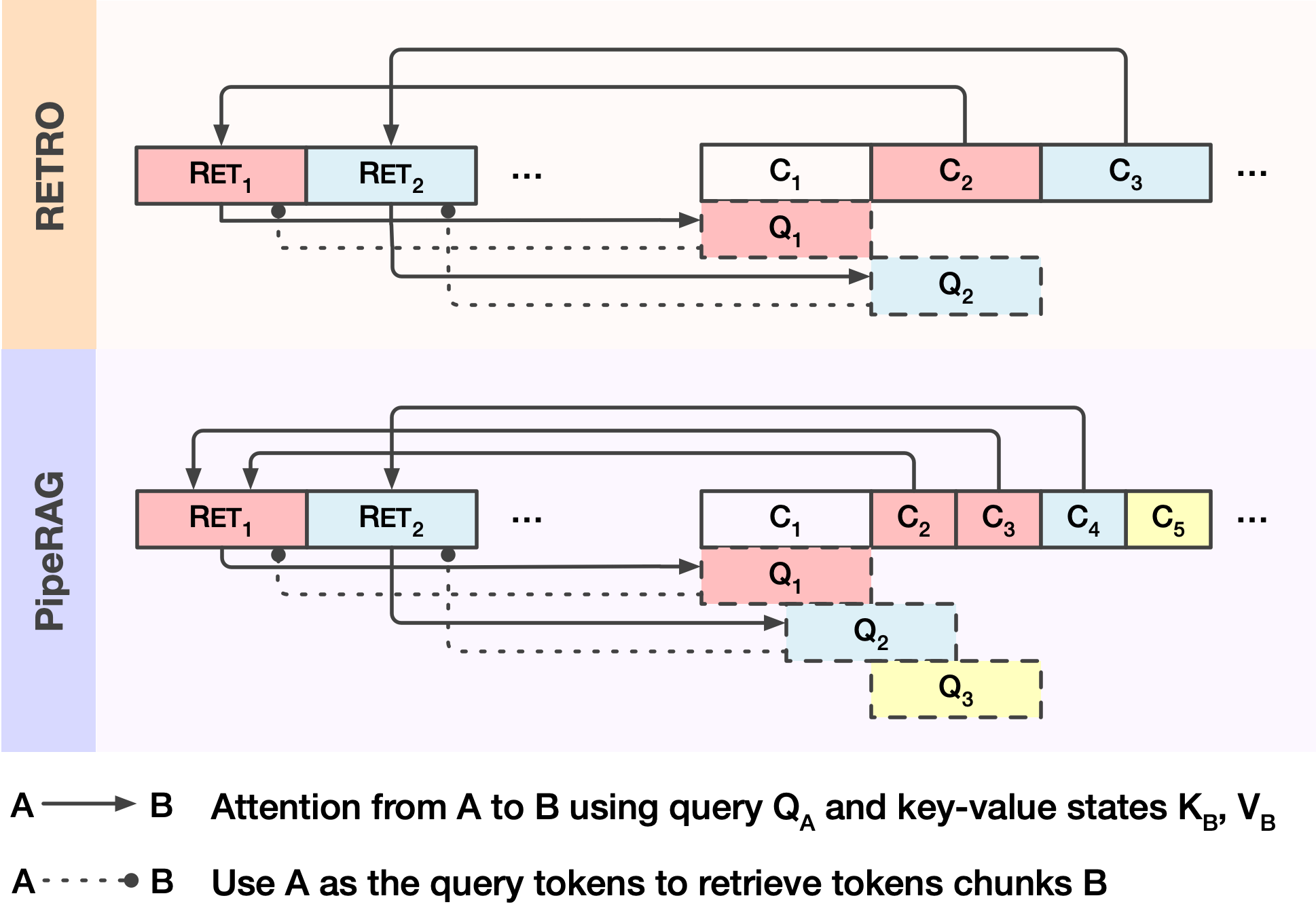}
  \vspace{-2em}
  \caption{Attention mechanisms and query windows in PipeRAG.}
  \vspace{-1em}
  \label{fig:attention}
\end{figure}

\textbf{S2: Flexible retrieval intervals.} 
\textsc{Retro} utilizes a fixed retrieval interval of \( m=64 \), aligning with the generation chunk size, database token chunk size, and query window size. 
However, the effectiveness of pipeline parallelism (S1) is maximized when the retrieval and inference subsystems have similar latencies --- generating \( m=64 \) tokens does not always consume similar time as one retrieval.

In order to improve the effectiveness of pipeline parallelism, PipeRAG supports alternative retrieval intervals \( m' \) and modifies \textsc{Retro}'s attention mechanism accordingly.
Here, \( m' \) remains constant during a single generation process but can vary from the default value of 64. When using shorter intervals, such as \( m'=32 \), the staleness of queries is also reduced (\( s=32 \), thereby improving the quality of the retrieved content to more closely resemble that obtained from a non-stale query.
Figure~\ref{fig:attention} illustrates the differences in retrievals and attention mechanisms between \textsc{Retro} and PipeRAG, taking \( m'=32 \) as an example. As shown in the figure, while a query \( Q_i \) still has a window size of \( m=64 \) tokens, the retrieval interval is halved. This necessitates adjustments in the attention regions to align with these modified intervals. For encoder-to-decoder attention, the attention is directed from the retrieved chunk to the query window whose position is different from that of \textsc{Retro}. For decoder-to-encoder attention, the generation of chunk \( C_{j+1} \) of length \( m' \) applies chunked cross-attention on \( \textsc{Ret}(Q_{j-1}) \).

\textbf{S3: Performance-model-driven retrievals.} 
PipeRAG has the potential to match the generation latency of LLMs that do not introduce retrievals, especially when the retrievals and inferences are completely overlapped in the pipeline.
However, achieving this ideal overlap is challenging because of the distinct performance characteristics of the retrieval and inference systems as introduced in O2 and O3. 

To address this, we propose a performance-model-driven retrieval system to automatically enable perfectly overlapped pipeline windows.
In this context, a performance model refers to any model (not limited to neural networks) designed to predict the performance characteristics of a system.
Specifically, the retrieval system takes the generation states as inputs and automatically adjusts the search space using performance models, ensuring that the retrieval latency can be hidden by the generation latency of the next token chunk. By maximizing the search space under the latency constraint, the retrieval quality is also maximized without incurring extra generation latency.

The inference performance can be modeled as follows. 
The time required to generate a token chunk is \( T_{C} = T_{\textsc{Enc}} + T_{\textsc{Dec}} \). The latency of encoder inference is related to the number of retrieved neighbors and the number of tokens per neighbor, while the decoder inference latency depends on the current sequence length and the chunk size (O2). 

On the other hand, the retrieval latency can be represented modeled as \( T_{\textsc{Ret}} = T_{Network} + T_{EncQuery} + T_{ScanIndex} + T_{ScanVec}\), encompassing the time spent on network communications, encoding the query tokens as vectors, scanning the vector index, and scanning a subset of database vectors. 
In this paper, we apply the widely-adopted IVF-PQ vector search algorithm~\cite{PQ} that combines a clustering-based inverted-file (IVF) index with product quantization (PQ). The IVF index clusters the database to \( nlist \) IVF lists. At query time, \( nprobe \) out of the \( nlist \) IVF lists are selected to scan (database vectors within the selected lists are compared to the query vectors). 

Given that the performance of both retrievals and inferences are related to hardware, we measure and model their performance on the deployment hardware. We record the time consumption of both encoder and decoder inferences with various input sequence lengths. For retrieval, we model the relationship between \( nprobe \) and search latency using linear regression, given that \( nprobe \) is approximately proportional to the number of scanned database vectors.

The retrieval system then leverages these performance models to predict the maximal search space, indicated by $nlist$, given the latency constraint for generating the next token chunk, ensuring that \( T_{\textsc{Ret}} \leq T(C) \). 
Since the \( T(C) \) can be easily obtained from the recorded performance numbers, we can then derive the maximal \( nprobe \) during the search based on the retrieval performance model.

While an alternative approach to achieve a perfectly overlapped pipeline is adjusting the retrieval intervals in the inference system, we rule out this option due to generalizability concerns. In future deployment scenarios, a retrieval system may serve multiple inference systems. Thus, the retrieval performance is impacted by the number of concurrent queries being processed. In this case, it could be challenging for the inference system to accurately predict the retrieval latency, as it lacks the information about the retrieval system's workload at the moment. Therefore, it is the retrieval system, instead of the inference system, that should be responsible for constructing a perfectly overlapped pipeline via performance modeling.








\begin{figure*}[t]
\begin{subfigure}
    \centering
    \includegraphics[width=0.32\linewidth]{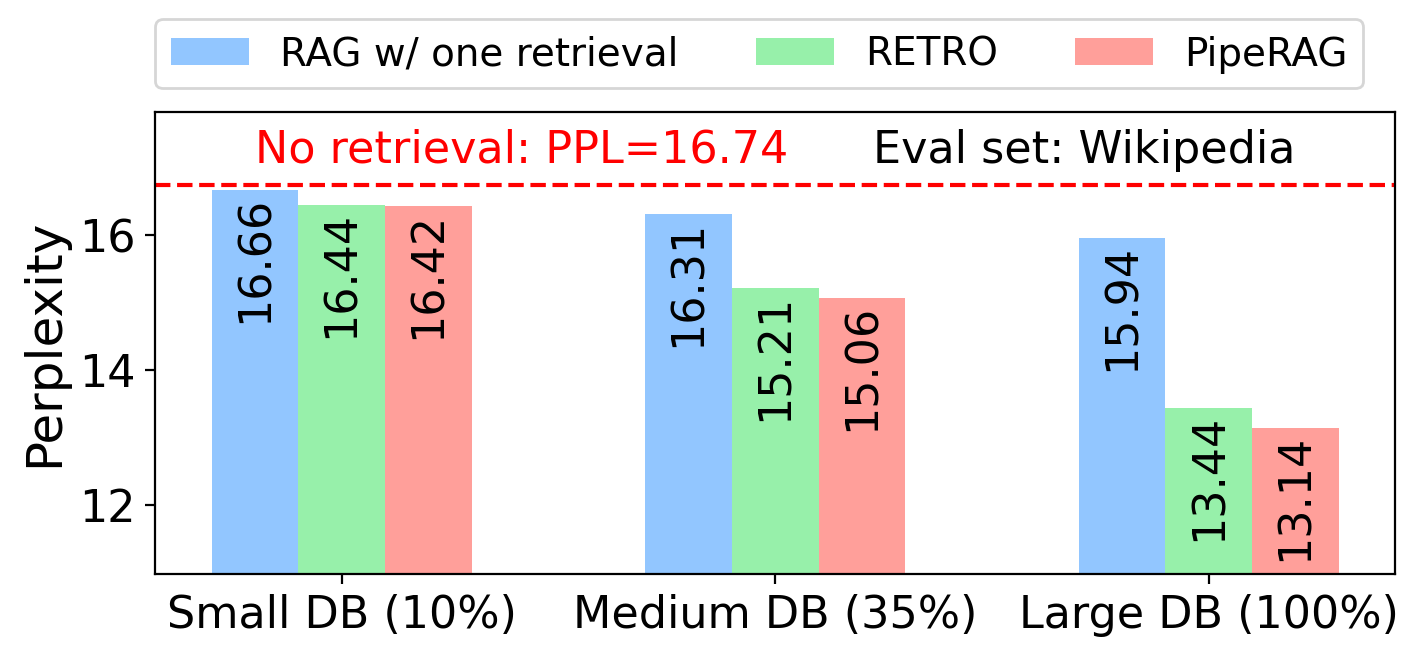}
\end{subfigure}
\hfill
\begin{subfigure}
    \centering
    \includegraphics[width=0.32\linewidth]{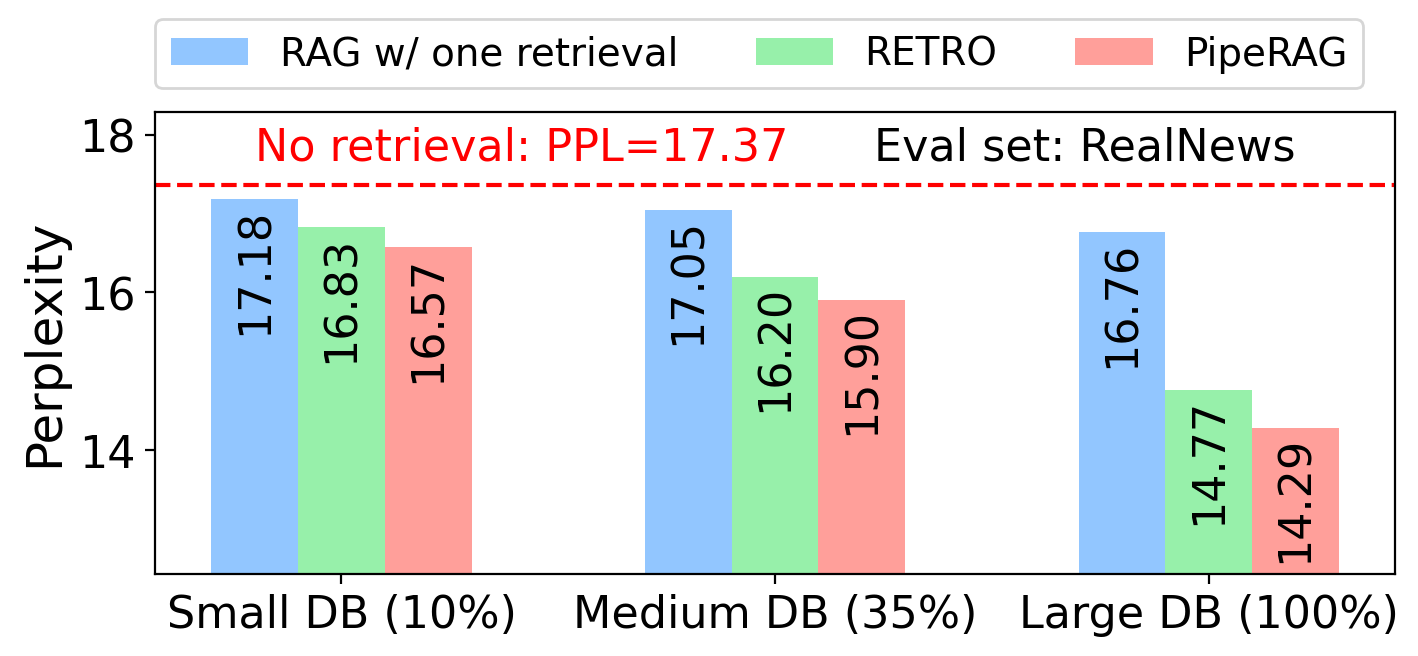}
\end{subfigure}
\hfill
\begin{subfigure}
    \centering
    \includegraphics[width=0.32\linewidth]{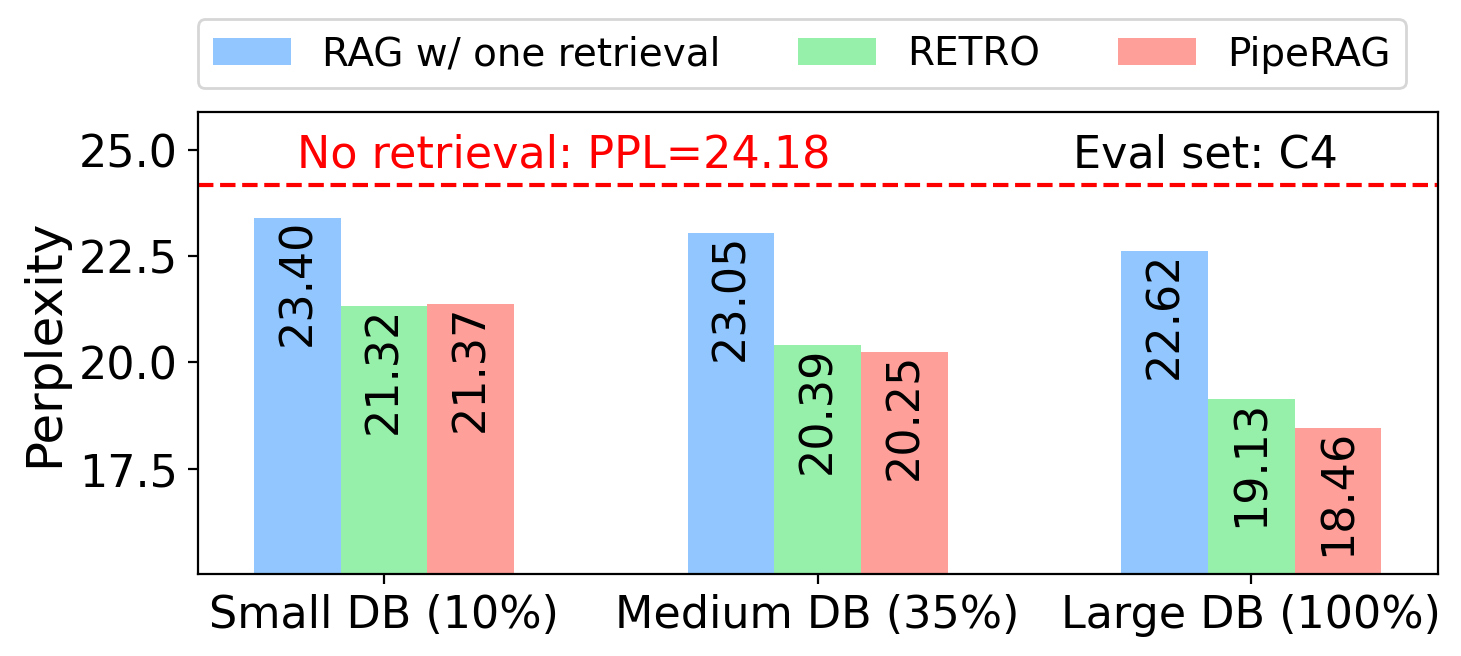}
\end{subfigure}
\hfill

  \vspace{-1em}
  \caption{The effect of database sizes and retrieval strategies on language modeling perplexity (lower perplexity means higher quality).}
  \label{fig:eval_dbsize}
\end{figure*}

\begin{figure*}
\begin{subfigure}
    \centering
    \includegraphics[width=0.32\linewidth]{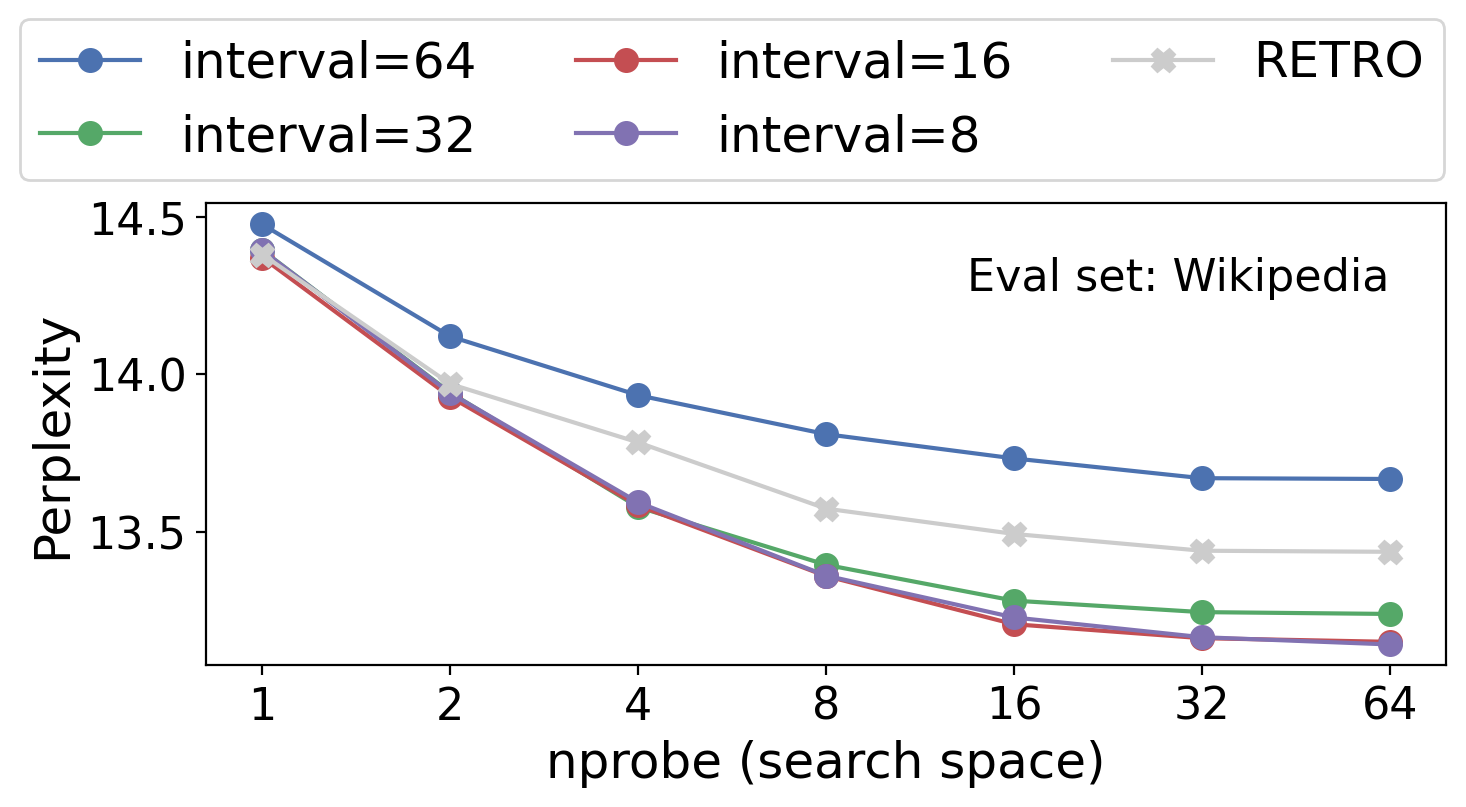}
\end{subfigure}
\hfill
\begin{subfigure}
    \centering
    \includegraphics[width=0.32\linewidth]{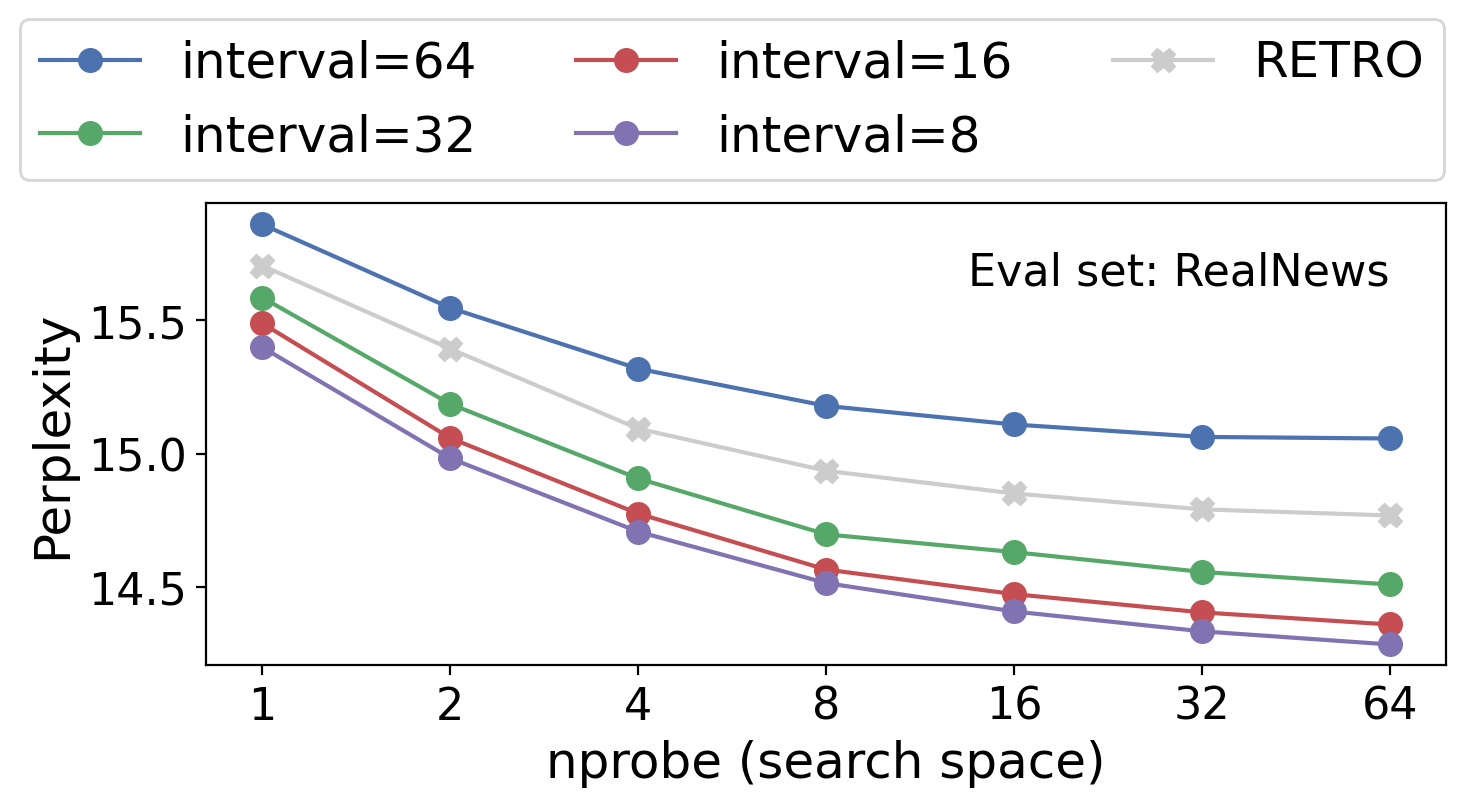}
\end{subfigure}
\hfill
\begin{subfigure}
    \centering
    \includegraphics[width=0.32\linewidth]{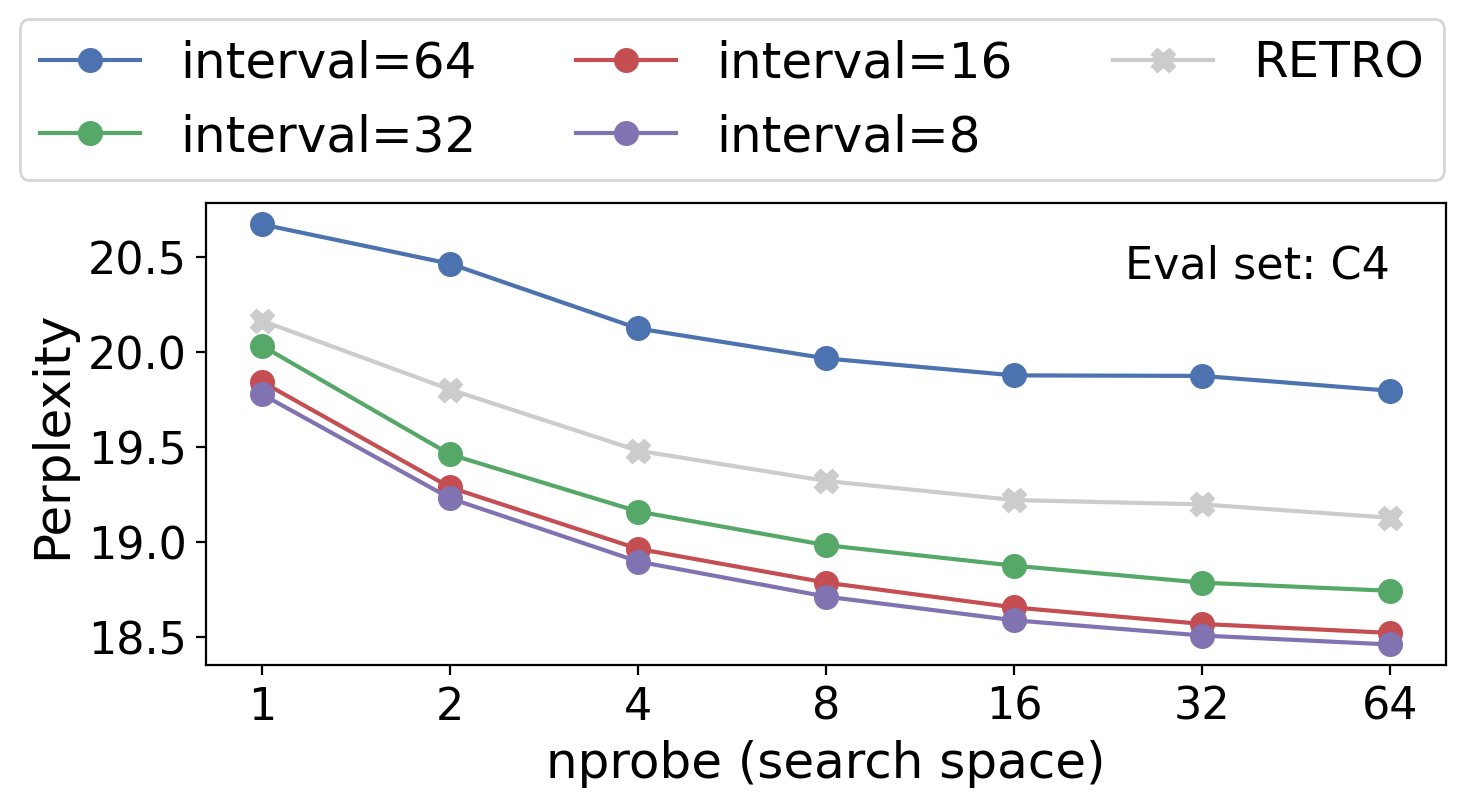}
\end{subfigure}
\hfill

  \vspace{-1em}
  \caption{Perplexity of retrieval-augmented generation when applying various retrieval intervals and search space configurations (\( nprobe \)).}
  \vspace{-1em}
  \label{fig:eval_ppl}
\end{figure*}

\section{Evaluation}
\label{sec:evaluation}

\subsection{Experimental Setup}

We briefly introduce our experimental setup below and leave more details in Appendix~\ref{sec:append_eval_setup}.



\textbf{Database.}
Our token database was constructed from the C4 corpus containing deduplicated English documents. Adhering to~\citet{borgeaud2022improving}, we segmented the documents into chunks of \( m=64 \) tokens, yielding a total of three billion chunks, and set the number of nearest neighbors per retrieval as \(k=2 \). Following~\citet{norlund2023generalization}, we transformed each token chunk into a 384-dimensional vector using a sentence transformer\cite{reimers2019sentence}.

\textbf{Model.}
We developed PipeRAG based on the \textsc{Retro} checkpoint with 582M parameters provided by~\citet{norlund2023generalization}, the only available pre-trained \textsc{Retro} model when we conducted the experiments.

\textbf{Evaluation Set.}
To evaluate language modeling quality, we used the Wikipedia dataset~\cite{wikipedia}, the RealNews subset of the C4 dataset, and C4's English document subset~\cite{dodge2021documenting, raffel2020exploring}.

\textbf{Software.}
For model inference, we adopted the ONNX runtime, which, according to our experiments, achieves 2 to 3 times speedup in latency over PyTorch. For retrieval, we used the Faiss library~\cite{johnson2019billion} and the IVF-PQ vector search algorithm. Communication between the inference and retrieval systems was managed via gRPC.

\textbf{Hardware.}
For model inference, we used an NVIDIA A100 GPU (40 GB), while the retrievals were conducted on a server equipped with Intel(R) Xeon(R) Platinum 8259CL CPUs @2.50GHz (48 cores) and 384 GB memory. 


\subsection{Perplexity Evaluation}

Figure~\ref{fig:eval_dbsize} shows the impact of various retrieval strategies across different database sizes. This comparison includes PipeRAG, \textsc{Retro}, retrieval-augmented generation with only one retrieval at the beginning of generation, and generation without retrieval. For the last two strategies, \textsc{Retro} still serves as the base model. 
As indicated in the figure, retrieval, especially on large databases, plays a crucial role in improving generation quality (lower perplexity is better). Across all evaluated datasets, generation without retrieval performs the worst, followed by only retrieving once, showing the effectiveness of periodic retrieval in \textsc{Retro}. Additionally, perplexity decreases as the dataset size increases, highlighting the importance of comprehensive content coverage in the databases. Notably, when pairing with the largest database, PipeRAG outperforms \textsc{Retro} in generation quality, as we will analyze in greater detail later on.


\textit{From now on, we report results in generation quality and performance based on the full (largest) database}, as using subsets significantly compromises generation quality.

Figure~\ref{fig:eval_ppl} compares the perplexity between PipeRAG and \textsc{Retro} across various retrieval configurations. We assess PipeRAG with different retrieval intervals, setting the search space through \( nprobe \), which represents the number of scanned vector lists per query in the IVF index. As shown in Figure~\ref{fig:eval_ppl}, both PipeRAG and \textsc{Retro} show reduced perplexity with an expanded search space, which leads to better search quality (O3).

\begin{tcolorbox}[
    enhanced,
    arc=2mm, 
    outer arc=2mm, 
    boxrule=0.8pt, 
    colframe=black, 
    colback=white, 
    boxsep=0pt, 
    drop shadow southeast, 
]

\textbf{Takeaway 1:} The quality of retrieval-augmented generation benefits from higher retrieval quality achieved by expanding the search space during vector search.
\end{tcolorbox}

\begin{figure*}[t]

\begin{subfigure}
    \centering
    \includegraphics[width=0.32\linewidth]{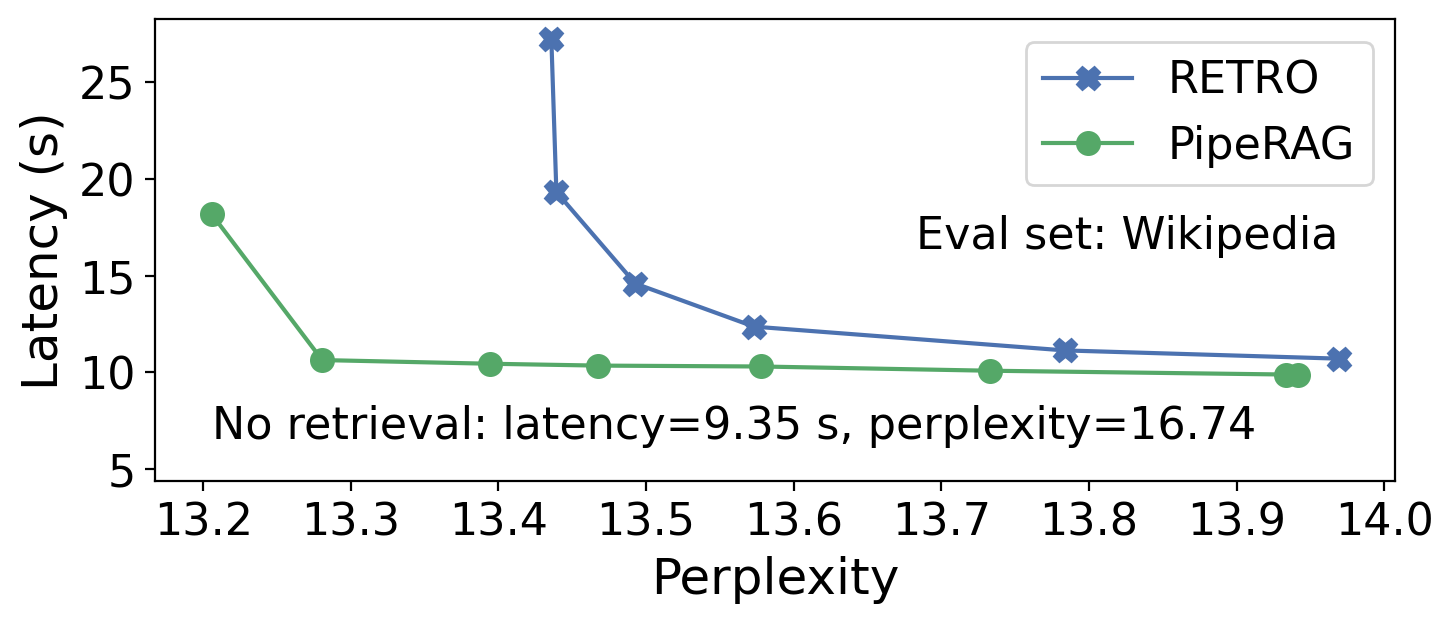}
\end{subfigure}
\hfill
\begin{subfigure}
    \centering
    \includegraphics[width=0.32\linewidth]{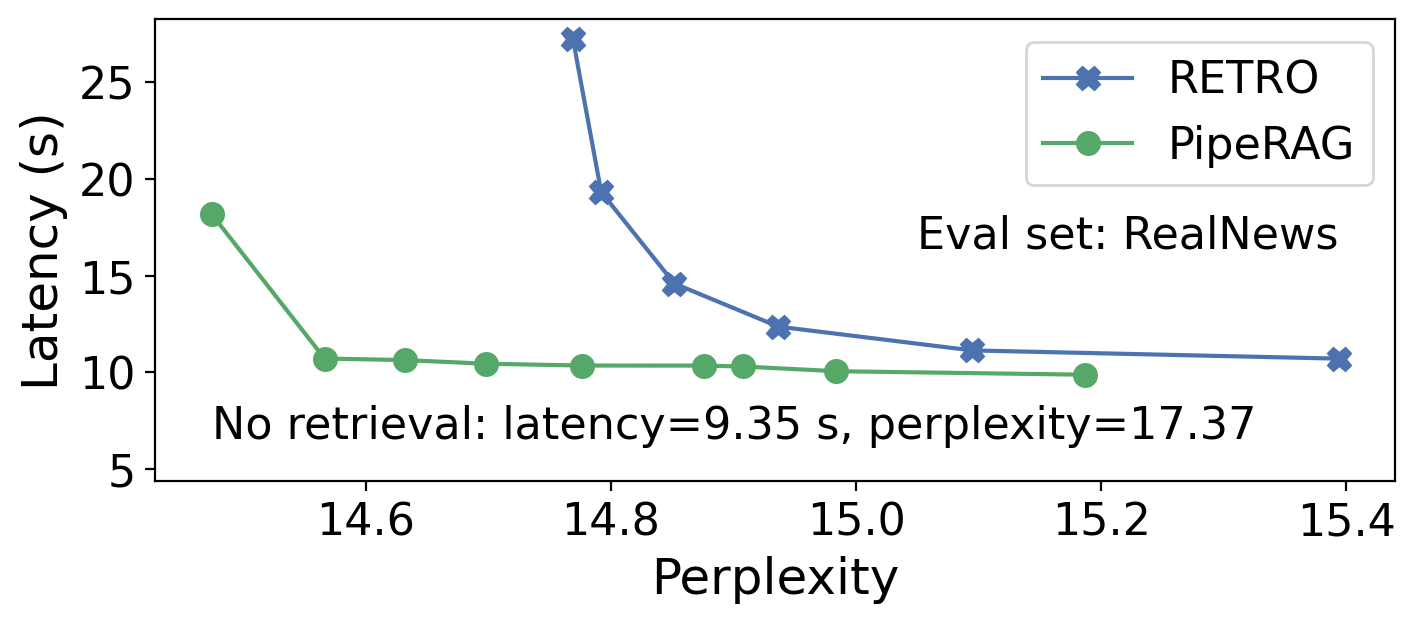}
\end{subfigure}
\hfill
\begin{subfigure}
    \centering
    \includegraphics[width=0.32\linewidth]{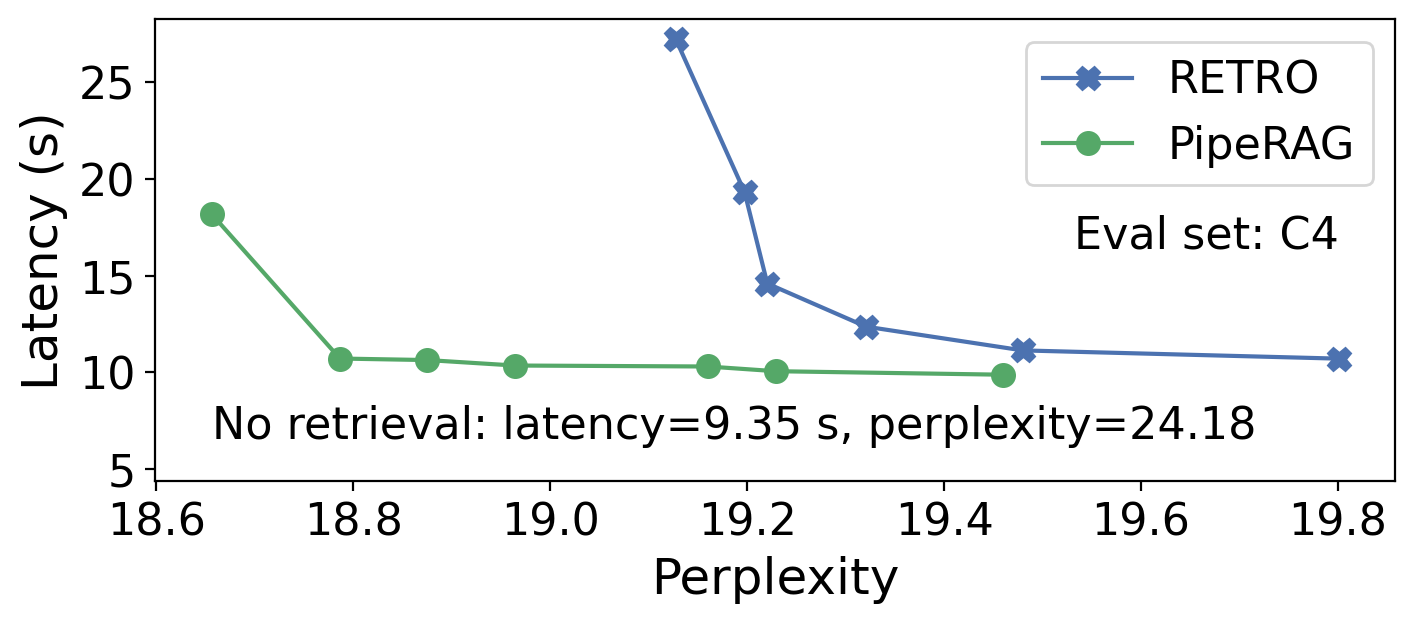}
\end{subfigure}

  \vspace{-1em}
  \caption{PipeRAG significantly outperforms \textsc{Retro} on the latency-perplexity Pareto frontier (lower latency and perplexity are better).}
  \vspace{-0.5em}
  \label{fig:eval_e2e}
\end{figure*}

\begin{table*}
\centering 

\caption{Performance-driven retrieval (S3) facilitates latency comparable to non-retrieval models while significantly reducing perplexity. Values in parentheses indicate the difference compared to the baseline model without retrieval (lower latency and perplexity are better).} 

\vspace{0.5em}
\scalebox{0.78}{
\begin{tabular}{
L{4em} 
M{5em} M{6em} M{14em} 
M{0em}
M{5em} M{6em} M{14em} 
}\toprule
\multirow{2}{*}{Eval Set} & \multicolumn{3}{c}{Latency (s)} & \phantom{}& \multicolumn{3}{c}{Perplexity} \\
\cmidrule{2-4} \cmidrule{6-8}
 & No retrieval & RETRO & Performance-driven retrieval (S3) &\phantom{}& No retrieval & RETRO & Performance-driven retrieval (S3) \\
\midrule
Wikipedia & 9.35 & 14.59 (+5.23) & 10.34 (\textbf{+0.99})  && 16.74 & 13.49 (-3.25) & 13.47 (\textbf{-3.28})    \\ 
RealNews & 9.35 & 12.36 (+3.00) & 10.58 (\textbf{+1.22})  && 17.37 & 14.94 (-2.43) & 14.87 (\textbf{-2.50})   \\ 
C4 & 9.35 & 11.13 (+1.78) & 10.58 (\textbf{+1.22})  && 24.18 & 19.48 (-4.70) & 19.36 (\textbf{-4.82})   \\ 

\bottomrule 
\end{tabular} }

\label{tab:dynamic_nprobe} 

\end{table*}

Furthermore, PipeRAG demonstrates superior generation quality over \textsc{Retro}, particularly when using shorter retrieval intervals of no more than 32 (Figure~\ref{fig:eval_ppl}). This advantage is attributed to PipeRAG's revised attention mechanism. Shorter intervals not only reduce query staleness (equivalent to the interval) but improve the content integration frequency, in contrast to \textsc{Retro} with a fixed interval of 64. The increased retrieval frequency in PipeRAG does not necessarily add to generation latency thanks to the pipeline parallelism, a point we will further elaborate on.

\begin{tcolorbox}[
    enhanced,
    arc=2mm, 
    outer arc=2mm, 
    boxrule=0.8pt, 
    colframe=black, 
    colback=white, 
    boxsep=0pt, 
    drop shadow southeast, 
]

\textbf{Takeaway 2:} PipeRAG can surpass \textsc{Retro} in generation quality when using shorter retrieval intervals backed by PipeRAG's attention mechanism.
\end{tcolorbox}

\subsection{RAG Efficiency: Performance-Quality Trade-offs}

In this section, we assess the efficiency of PipeRAG. Our primary performance metric is the end-to-end latency to generate a 1024-token sequence, which we reported by taking the median latency of five individual runs. 


Figure~\ref{fig:eval_e2e} compares the Pareto frontiers of the performance-quality (latency-perplexity) trade-offs between PipeRAG and \textsc{Retro}.
For \textsc{Retro}, we manipulate the search space by tuning \( nprobe \).
For PipeRAG, we explore a range of retrieval intervals in conjunction with either a fixed search space or the performance-model-driven search space selection (S3).  
Across all datasets, the Pareto frontier of PipeRAG demonstrates significant advantages over \textsc{Retro}, as shown in Figure~\ref{fig:eval_e2e}. For example, PipeRAG can attain up to a 2.6$\times$ reduction in latency while maintaining or reducing perplexity relative to \textsc{Retro}; alternatively, under the same latency constraint, PipeRAG can lower perplexity by as much as 0.93 points compared to \textsc{Retro}.

\begin{tcolorbox}[
    enhanced,
    arc=2mm, 
    outer arc=2mm, 
    boxrule=0.8pt, 
    colframe=black, 
    colback=white, 
    boxsep=0pt, 
    drop shadow southeast, 
]
\textbf{Takeaway 3:} PipeRAG shows impressive efficiency, achieving up to 2.6$\times$ speedup in latency over \textsc{Retro} without compromising generation quality.
\end{tcolorbox}

Table~\ref{tab:dynamic_nprobe} demonstrates the effectiveness of the proposed performance-model-driven retrieval system. The objective of the performance model is to dynamically maximize search quality while minimizing additional performance costs.
To evaluate this, we compare the generation latency and quality of PipeRAG applying performance-model-driven retrievals to that of \textsc{Retro} as well as the same base \textsc{Retro} model without invoking retrievals.
As shown in Table~\ref{tab:dynamic_nprobe}, PipeRAG achieves a notable reduction in perplexity (2.50$\sim$4.82) with a minor increase in performance overhead (merely 10.6\%$\sim$13.2\% in latency overhead), outperformance \textsc{Retro} in both latency and perplexity. 
This slight increase in latency is attributed to the extra computational workload of the cross-attention mechanism when integrating the retrieved content from the encoder. 


\begin{tcolorbox}[
    enhanced,
    arc=2mm, 
    outer arc=2mm, 
    boxrule=0.8pt, 
    colframe=black, 
    colback=white, 
    boxsep=0pt, 
    drop shadow southeast, 
]

\textbf{Takeaway 4:} 
Leveraging the performance-model-driven retrieval system, PipeRAG can achieve comparable latency to models without retrievals while significantly improving generation quality. 
\end{tcolorbox}

\begin{figure}
	\centering
  \includegraphics[width=1.0\linewidth]{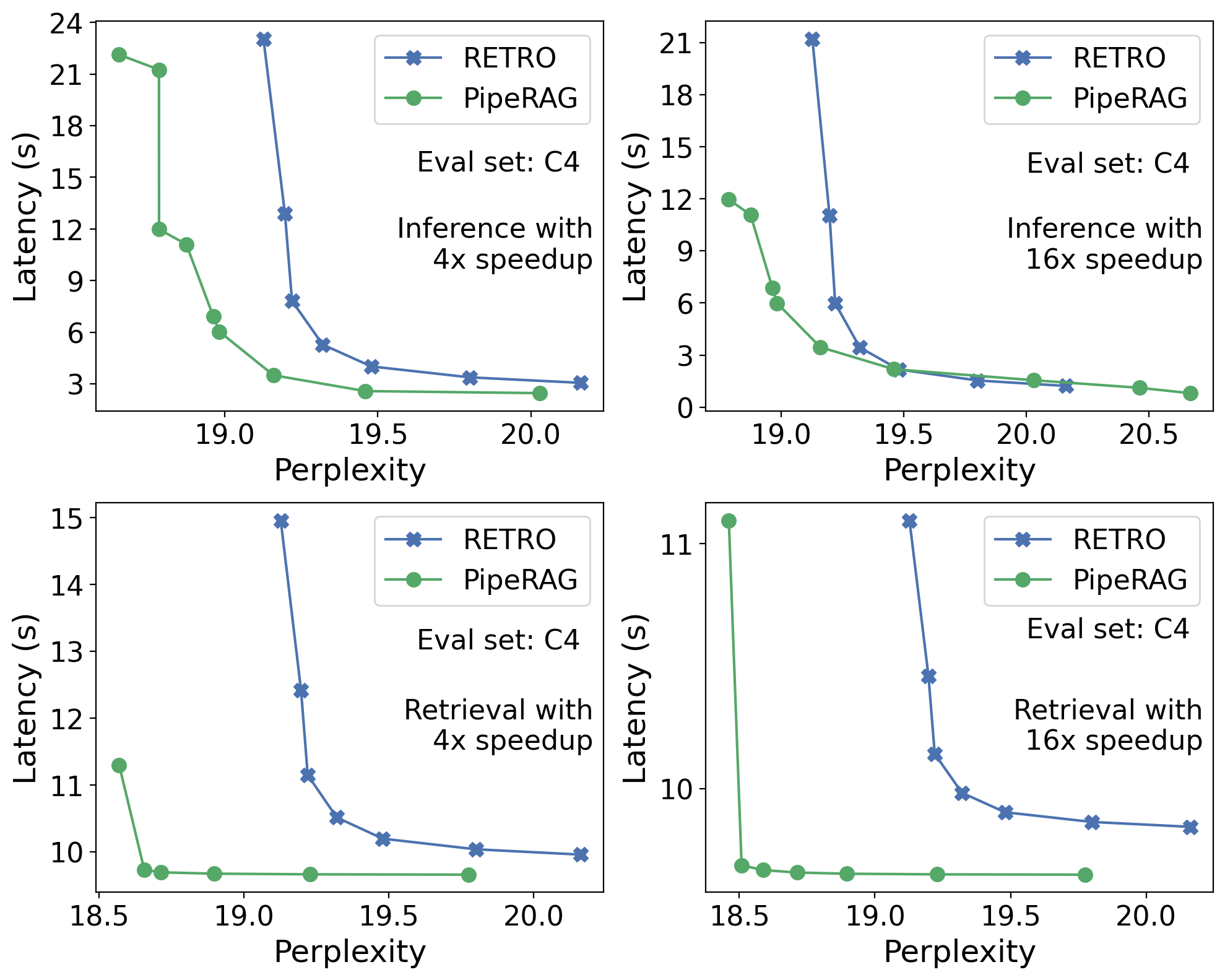}
  \vspace{-1.5em}
  \caption{Trends in PipeRAG efficiency when deployed on future hardware that enables faster retrieval or inference.}
  \vspace{-1em}
  \label{fig:different_performance}
\end{figure}


Figure~\ref{fig:different_performance} illustrates the projected performance trends of PipeRAG across a range of system and hardware configurations. Considering the rapid advancements in hardware accelerators, we expect shifts in performance of both retrieval and inference systems over years. To analyze PipeRAG's effectiveness on future hardware, we model the latency of PipeRAG and \textsc{Retro} when using faster retrieval or inference systems, with the methodology described in Appendix~\ref{sec:performance_trend}. 
The first row of Figure~\ref{fig:different_performance} demonstrates the generation latency when the inference system becomes 4$\times$ and 16$\times$ faster, while the second row examines the effects of accelerated retrieval. Across all scenarios, PipeRAG achieves superior efficiency compared to \textsc{Retro}. When either system experiences an order of magnitude speedup (e.g., 16$\times$), however, the benefits of applying PipeRAG become less significant. This trend aligns with our expectations, as the effectiveness of pipeline parallelism peaks when both system components have comparable latencies and diminishes when one component significantly outpaces the other.

\begin{tcolorbox}[
    enhanced,
    arc=2mm, 
    outer arc=2mm, 
    boxrule=0.8pt, 
    colframe=black, 
    colback=white, 
    boxsep=0pt, 
    drop shadow southeast, 
]

\textbf{Takeaway 5:} PipeRAG outperforms \textsc{Retro} in efficiency across different hardware, though the extent of improvements depends on sub-system performance.
\end{tcolorbox}

\subsection{Ablation Study}

Since PipeRAG not only introduces pipeline parallelism but also modifies \textsc{Retro}'s attention mechanism to maximize the effectiveness of pipelining, it is natural to ask how a baseline model would perform if it integrates the same attention mechanism. To illustrate the effectiveness of pipeline parallelism itself, we compare PipeRAG with an enhanced variant of \textsc{Retro}, named \textsc{Retro+}, which also supports flexible retrieval intervals by integrating PipeRAG's attention mechanism.

\begin{figure}[t]

\begin{subfigure}
    \centering
    \includegraphics[height=6.2em]{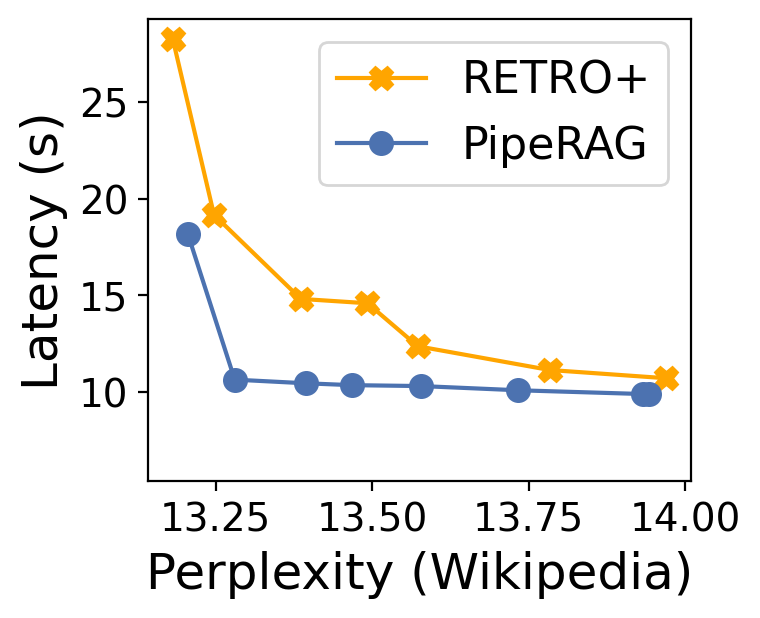}
\end{subfigure}
\hfill
\begin{subfigure}
    \centering
    \includegraphics[height=6.2em]{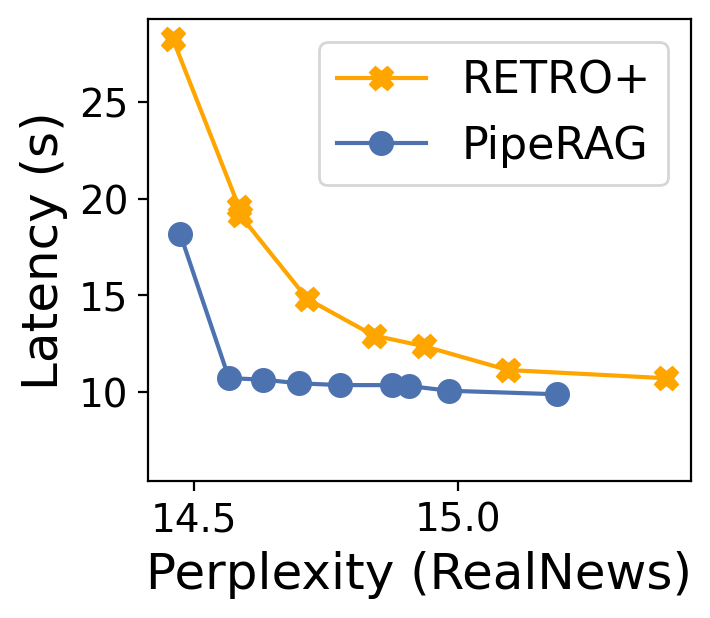}
\end{subfigure}
\hfill
\begin{subfigure}
    \centering
    \includegraphics[height=6.2em]{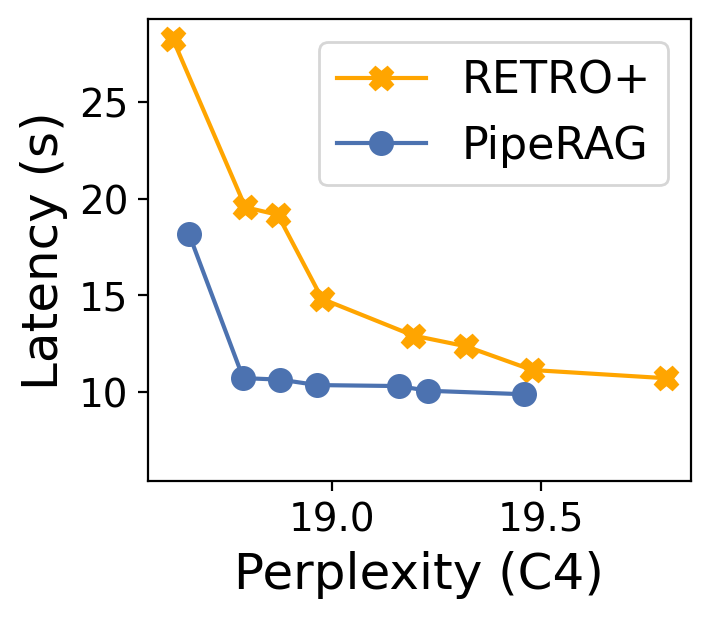}
\end{subfigure}

  \vspace{-1em}
  \caption{Even if the baseline model supports flexible retrieval intervals (\textsc{Retro+}), PipeRAG still significantly outperforms it in efficiency thanks to the proposed pipeline parallelism.}
  \vspace{-1em}
  \label{fig:eval_e2e_retro_flexible_interval}
\end{figure}

Figure~\ref{fig:eval_e2e_retro_flexible_interval} compares the performance-quality Pareto-frontier between PipeRAG and \textsc{Retro+}. 
Both models use retrieval intervals ranging from 8 to 64.
While \textsc{Retro+}, benefiting from flexible intervals, matches PipeRAG in perplexity, PipeRAG consistently achieves lower latency given the same perplexity. This is attributed to the proposed pipeline parallelism: PipeRAG effectively hides the retrieval latencies by overlapping them with generation latencies, whereas for \textsc{Retro+}, more frequent retrievals lead to increased total generation latency. More detailed comparisons between PipeRAG and \textsc{Retro+} under identical retrieval intervals (corresponding to the same number of database requests) can be found in Appendix~\ref{sec:append_more_results}.


\begin{tcolorbox}[
    enhanced,
    arc=2mm, 
    outer arc=2mm, 
    boxrule=0.8pt, 
    colframe=black, 
    colback=white, 
    boxsep=0pt, 
    drop shadow southeast, 
]

\textbf{Takeaway 6:} Pipeline parallelism is essential to achieve superior RAG efficiency, as PipeRAG outperforms \textsc{Retro+} that supports flexible retrieval intervals using PipeRAG's attention mechanism.
\end{tcolorbox}

\section{Related Work}

To the best of our knowledge, PipeRAG represents the first endeavor to enhance RAG efficiency through an in-depth algorithm-system co-design, diverging from existing RAG research that mainly focuses on improving generation quality. We now briefly introduce these related works.

Since knowledge is primarily retrieved rather than encoded in the LLM's parameters, RALMs, even with LLMs of one to two orders of magnitude fewer parameters, can achieve superior or comparable performance to conventional LLMs on various natural language processing (NLP) tasks~\cite{lewis2020pre, izacard2022few, komeili2021internet, guu2020retrieval}. While the generation may only involve a single passage retrieval at the beginning~\cite{lewis2020retrieval, izacard2020leveraging, sachan2021end}, the generated sequence may gradually diverge from the initially retrieved contents as the sequence grows longer. Thus, a more generaral RAG approach involves multiple retrievals during text generation to improve token generation quality~\cite{ram2023context, borgeaud2022improving}.

Another line of RAG research emphasizes token-level retrievals, exemplified by kNN-LM~\cite{khandelwal2019generalization} and subsequent works~\cite{khandelwal2020nearest, meng2021fast, xu2023nearest}. In these models, during each token generation step, the hidden state of the last layer is used as a query to retrieve contextually similar passages as well as their subsequent tokens (with a retrieval interval of one). The next token of the current context is then predicted by interpolating the model's next-token probability distribution with that of the retrieved contents. There are also arguments suggesting that token-level content integration may not be as effective as integrating longer passages~\cite{wang2023knn}.

\section{Conclusion}
\label{sec:conclusion}

Retrieval-augmented generation presents both opportunities and efficiency challenges, due to the significant overheads when retrieving from large databases. 
We propose PipeRAG, a novel RAG approach that improves generation efficiency by adopting pipeline parallelism, allowing flexible retrieval intervals, and dynamically adjusting retrieval quality via performance modeling. PipeRAG achieves up to 2.6$\times$ speedup over \textsc{Retro} without compromising generation quality. This not only establishes a solid foundation for integrating pipeline parallelism in future RAG systems but also showcasing future research opportunities in optimizing RAG through algorithm-system co-design.


\newpage
\section*{Impact Statements}
\label{sec:impact}

This paper focuses on enhancing the system efficiency of retrieval-augmented generation, aiming to reduce both energy consumption and carbon emissions during large-scale LLM inference. As our work does not involve training new models, we anticipate minimal ethical concerns or adverse societal impacts.

\bibliography{ref}
\bibliographystyle{icml2023}

\newpage
\appendix
\onecolumn

\section{A Motivating Example of Periodic Retrievals}
\label{sec:append_more_background}

In this section, we present a concrete example demonstrating the effectiveness of \textit{periodic retrievals} during sequence generation, a strategy that has been proven to significantly enhance the quality of language modeling~\cite{borgeaud2022improving, ram2023context, norlund2023generalization}.

\begin{figure*}[h]
	\centering
  \includegraphics[width=1.0\linewidth]{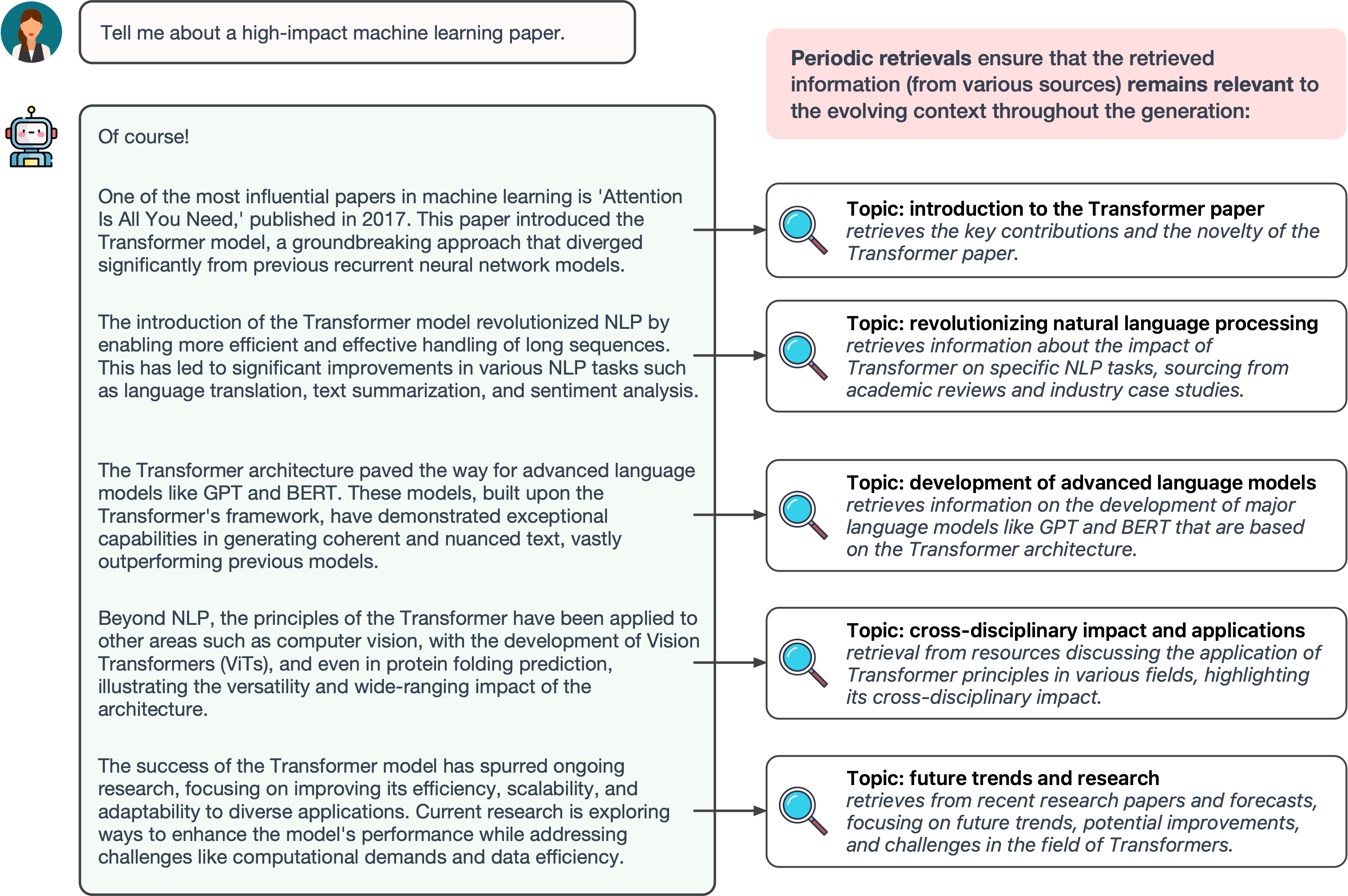}
  \vspace{-1em}
  \caption{A motivating example of utilizing periodic retrievals during sequence generation.}
  \label{fig:example_periodic_retrievals}
\end{figure*}

Figure~\ref{fig:example_periodic_retrievals} illustrates the example, wherein the model is asked to describe a high-impact machine learning paper. In crafting its response, the model uses the Transformer neural network~\cite{vaswani2017attention} as the target paper, covering several aspects related to the paper. The narrative evolves from a brief introduction of the model, through its impacts on various natural language processing tasks, to its influence on subsequent research, its cross-disciplinary applications, and ultimately, to emerging trends in research. Given these shifts in topic, the content initially retrieved about the Transformer architecture might lose relevance in the context of discussing future research trends. Therefore, periodic retrievals, in this instance, are vital to ensure that the retrieved content remains pertinent to the current context of generation.

\vspace{1em}


\section{Detailed Evaluation Setup}
\label{sec:append_eval_setup}

\textbf{Datasets.}
We constructed the token database from the C4 dataset, using deduplicated English documents. We did not choose the Pile dataset used in previous works~\cite{borgeaud2022improving} due to its current copyright issues.
By segmenting these documents into chunks of \( m=64 \) tokens, we generated a total of three billion chunks.
Subsequently, each chunk was converted into a 384-dimensional vector using a sentence transformer~\cite{reimers2019sentence} checkpoint \textit{all-MiniLM-L6-v2}.

\textbf{Software.}
Our implementation of the PipeRAG model is based on a \textsc{Retro} baseline obtained from~\cite{norlund2023generalization}, which is built on top of PyTorch.  To enhance inference performance, we supported the caching of key-value states in the transformer and converted the model to ONNX format, enabling model inference by ONNX runtime. With the above optimizations, the inference latency on GPU is improved by around 3$\times$ over the original Pytorch implementation. We maintained the fp32 (32-bit floating point) precision of the model.

For the retrieval system, we used the Faiss library~\cite{johnson2019billion}, which is known for its efficient product-quantization-based vector search implementation. 
We adopted the IVF-PQ vector search algorithm, setting the number of IVF list centroids to \( nlist=16384 \) and quantizing each 384-dimensional vector into 64 bytes of PQ code. 
During retrievals, we set the number of nearest neighbors as \(k=2 \).

The communication between the inference and retrieval systems was managed via the gRPC library.

\textbf{Hardware.}
We used two separate platforms for inference and retrievals. For model inference, we utilized an NVIDIA A100 GPU (40 GB). The retrieval process was handled by a server with substantial memory capacity to accommodate the large encoded dataset. The server was equipped with dual-socket Intel(R) Xeon(R) Platinum 8259CL CPUs @2.50GHz (48 cores and 96 threads) and 384 GB memory. 
The retrieval and inference servers were interconnected through a network, with a round-trip time (RTT) of around 1 ms.




\section{Performance Trends on Evolving Hardware}
\label{sec:performance_trend}

In this section, we begin by enumerating the factors that influence retrieval and inference performance. We then introduce the performance modeling methodology employed in Section~\ref{sec:evaluation}, which projects PipeRAG's efficiency on future hardware configurations.

\subsection{Factors Influencing Retrieval and Inference Performance}

\textbf{Retrieval performance} depends on the following factors:

\begin{itemize}
    \item \textbf{Hardware.} The memory bandwidth and computational capacity of the hardware used for retrieval are key factors influencing performance. It is worth noticing that there are emerging hardware accelerators that are specialized for retrievals~\cite{jiang2023co} and integrated into RAG systems~\cite{jiang2023chameleon}, offering impressive retrieval performance as well as cost efficiency.
    \item \textbf{Document numbers.} The total number of documents, along with encoding granularity as introduced below, determines the vector count in the database.
    \item \textbf{Encoding granularity.} Documents can be encoded in various granularities by LLMs, ranging from one vector per document~\cite{huang2013learning, karpukhin2020dense} to one vector per passage~\cite{dai2019deeper, reimers2019sentence} or even per token~\cite{khattab2020colbert, santhanam2021colbertv2}.
    \item \textbf{Dimensionality.} The dimensionality of the database vectors, as well as the compression ratio when employing product quantization, are critical to retrieval performance.
    \item \textbf{Indexes.} The selection of indexes, such as IVF or graph-based ones, and their parameter configurations are crucial for retrieval efficiency.
    \item \textbf{Reranking.} Optionally, the retrieved content can be reranked using LLMs, which often yields better ranking quality than relying solely on vector similarity~\cite{nogueira2019passage}.
\end{itemize}

\textbf{LLM inference performance} is influenced by the following factors:

\begin{itemize}
    \item \textbf{Hardware.} The performance of inference is heavily dependent on the hardware, particularly its memory bandwidth and computational capacity. LLM accelerators such as GPUs are evolving rapidly in these metrics. 
    \item \textbf{Software.} The choice of software for inference also plays a significant role. For instance, PyTorch's eager execution mode might not fully exploit hardware accelerators due to the slow execution speed of Python programs. In such cases, software overhead could exceed the GPU kernel execution time.
    \item \textbf{Quantization}. Quantizing models to lower precisions can markedly reduce inference time, thanks to reduced memory footprint and bandwidth usage. For instance, converting models to 3-bit precision can lead to a 3$\sim$5$\times$ speedup compared to 16-bit floating point formats~\cite{frantar2022gptq}.
    \item \textbf{Sparsity}. Techniques like mixture-of-experts allow for scaling LLMs without proportionate increases in computational costs~\cite{fedus2022switch, du2022glam}, because only a small subset of neurons are activated during inference. 
\end{itemize}

\subsection{Performance Modeling for Future Hardware}

To estimate PipeRAG's efficiency on future hardware, we model its performance using hypothetical hardware with enhanced inference and/or retrieval performance. We included the modeled performance in Section~\ref{sec:evaluation},  with a detailed explanation of our performance modeling approach provided here.

For \textsc{Retro}, the end-to-end generation latency is the sum of inference and retrieval time. In PipeRAG, due to the parallelism, the latency for generating a chunk of tokens is determined by the maximum value of the inference and retrieval latency of that chunk, except for the first chunk where the pipeline is not yet active (see Figure~\ref{fig:overview}).

We then input the measured performance of inference and retrievals into the performance model. This allows us to simulate performance scaling, such as a 4$\times$ improvement in retrieval or a 16$\times$ enhancement in inference. The result generation latency as well as the respective conclusions are included in Section~\ref{sec:evaluation}. The model's accuracy is then verified by comparing these projected results against actual experimental data, with deviations found to be within a reasonable range (the median difference is only 5.7\%).


\section{Additional Experimental Results}
\label{sec:append_more_results}

In this section, we include additional experimental results to further illustrate the effectiveness of PipeRAG.
First, we demonstrate the fundamental applicability of pipeline parallelism by illustrating the effectiveness of prefetching content with stale queries. Second, we show the advantages of PipeRAG over a modified version of RETRO, which, similar to PipeRAG, supports flexible retrieval intervals, highlighting the benefits of pipeline parallelism.

\subsection{The Effectiveness of Retrievals using Stale Queries}

We investigate the fundamental applicability of prefetching content using stale queries. For this purpose, we compare the \( k=1 \) nearest neighbors retrieved by non-stale queries in our evaluation set with their staleness versions. Same as Section~\ref{sec:evaluation}, we use the largest C4 database, which consists of three billion token chunks, and set \( nprobe=64 \) to ensure high retrieval quality. We then employ the \textit{msmarco-bert-base-dot-v5} checkpoint from sentence transformers~\cite{reimers2019sentence} to evaluate the cosine similarity between contents retrieved by stale and non-stale queries.

Table~\ref{tab:stale_query_results_similarity} presents the retrieval quality using stale queries. Here, we use different degrees of staleness, ranging from 1 token to 64 tokens, while maintaining a consistent retrieval interval of \( m=64 \). The results indicate that, despite the staleness, the retrieved content closely resembles that obtained through non-stale queries, with around 90\% cosine similarity across datasets. As expected, this similarity shows a gradual decline as the staleness increases.

\begin{table*}[h]
\centering 

\caption{Cosine similarity between content retrieved by stale and non-stale queries. The results indicate that stale queries are still highly effective in identifying relevant token chunks from the database.} 

\vspace{1em}
\scalebox{0.85}{
\begin{tabular}{
L{4em} M{5em} M{0em}
M{5em} M{5em} M{5em} M{5em} M{5em} M{5em} M{5em} 
}\toprule
& \multirow{2}{*}{No staleness} & \phantom{}& \multicolumn{7}{c}{Staleness (number of stale tokens in the query)} \\
 \cmidrule{4-10}
 &&& 1 & 2 & 4 & 8 & 16 & 32 & 64 \\
\midrule
Wikipedia & 1.0000 && 0.9262 & 0.9204 & 0.9138 & 0.9062 & 0.8990 & 0.8921 & 0.8875 \\
RealNews & 1.0000 && 0.9219 & 0.9147 & 0.9073 & 0.8996 & 0.8925 & 0.8850 & 0.8794 \\
C4 & 1.0000 && 0.9323 & 0.9263 & 0.9193 & 0.9127 & 0.9052 & 0.8980 & 0.8929 \\

\bottomrule 
\end{tabular} }
\label{tab:stale_query_results_similarity}
\end{table*}

\subsection{PipeRAG versus Baseline Model that Supports Flexible Retrieval Intervals}

To further show the efficiency gains of pipeline parallelism, we also compare PipeRAG with a modified version of \textsc{Retro}, termed \textsc{Retro+}, which also supports flexible retrieval intervals as PipeRAG. Here, we extend the results in Figure~\ref{fig:eval_e2e_retro_flexible_interval} 
 presented in Section~\ref{sec:evaluation}.

\begin{figure*}[h!]

\begin{subfigure}
    \centering
    \includegraphics[width=1.0\linewidth]{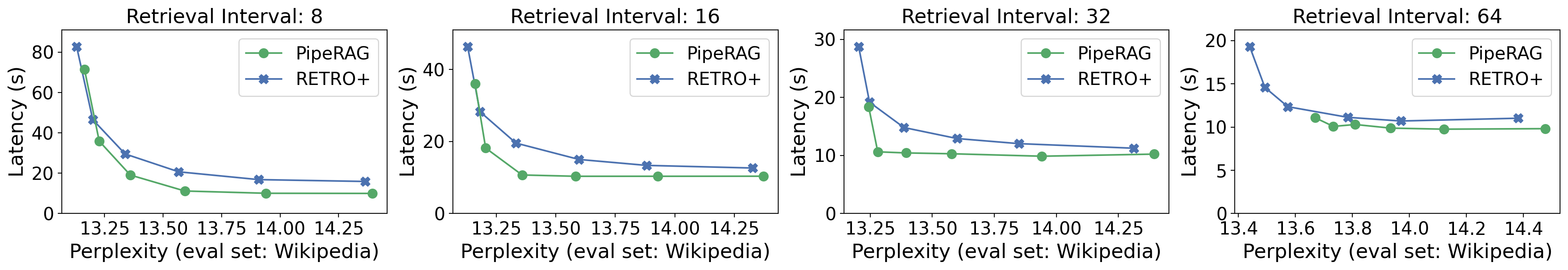}
\end{subfigure}
\hfill
\begin{subfigure}
    \centering
    \includegraphics[width=1.0\linewidth]{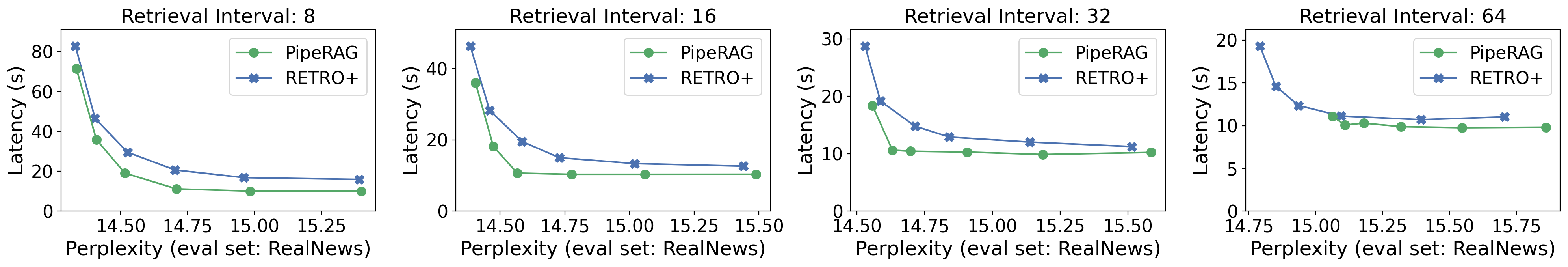}
\end{subfigure}
\hfill
\begin{subfigure}
    \centering
    \includegraphics[width=1.0\linewidth]{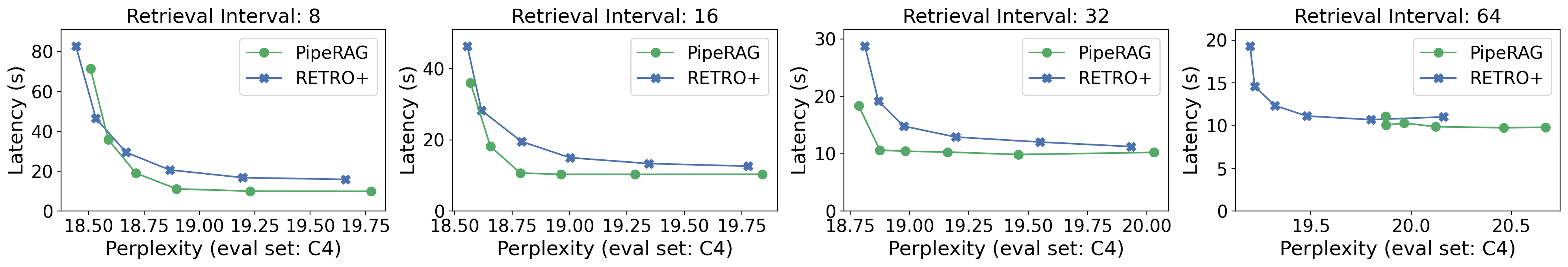}
\end{subfigure}

  \vspace{-1em}
  \caption{The latency-perplexity comparison between PipeRAG versus \textsc{Retro} given the same retrieval intervals.}
  \label{fig:eval_same_interval}
\end{figure*}

Figure~\ref{fig:eval_same_interval} presents a performance-quality comparison between PipeRAG and \textsc{Retro+} under identical retrieval intervals (corresponding to the same number of database requests). For most retrieval intervals, ranging from 8 to 32, PipeRAG demonstrates superior efficiency compared to \textsc{Retro+}. When the staleness is high (at a retrieval interval of 64), \textsc{Retro+} has the potential to outperform PipeRAG in scenarios of low perplexity, attributable to the effects of high query staleness.




\section{Broader Applicability of PipeRAG}
\label{sec:discussion}

\textit{The idea of improving RAG efficiency through pipeline parallelism is broadly applicable across various RAG configurations, as long as they include periodic retrievals.} 
In this paper, we have focused on improving RAG efficiency based on the \textsc{Retro} model and evaluated generation performance using specific hardware and software setups described in Section~\ref{sec:evaluation}. 
In the future, RAG can evolve in several ways: models may adopt a decoder-only transformer architecture~\cite{radford2018improving, brown2020language} although the high cost of periodically appending the retrieved content has to be addressed~\cite{ram2023context, jiang2023active}; retrieval engines could incorporate LLM-based or BM25-based result reranking~\cite{nogueira2019passage, macavaney2019cedr, doostmohammadi2023surface}, instead of solely relying on vector-level similarity; and hardware may evolve to include dedicated retrieval accelerators~\cite{jiang2023co, jiang2023chameleon}. 
However, regardless of these potential advancements in algorithms and hardware (detailed in Appendix~\ref{sec:performance_trend}), the dependencies between retrievals and inferences in RAG systems --- especially when retrievals are periodic --- remains a \textit{fundamental} obstacle to fully leveraging hardware resources and achieving maximal inference efficiency. Thus, whenever the time consumption of one retrieval and multiple steps of inferences are on a similar scale, pipeline parallelism by \textit{prefetching} content from databases should be a great option to improve generation efficiency. 

\textit{Prefetching content from databases using stale queries is applicable regardless of the specific models used for generation.} To demonstrate this, we show that using a stale query window can retrieve content very similar to that obtained via a regular query window, with detailed results included in Appendix~\ref{sec:append_more_results}. 
These findings address a potential limitation in our evaluation, as our experimentation with PipeRAG was conducted using the \textsc{Retro} checkpoint provided by~\cite{norlund2023generalization}, which was the only available \textsc{Retro} checkpoint at the time of our research.

\end{document}